\documentclass[twoside,onecolumn,draft]{IEEEtran}
\usepackage{cite}

\ifCLASSINFOpdf
\usepackage{amsmath}
\usepackage[pdftex]{graphicx}
\usepackage[vlined, ruled, boxed]{algorithm2e}
\usepackage{subfigure}
\usepackage{graphicx}
\usepackage{url}
\usepackage{epstopdf}
\else
  \usepackage[dvips,final]{graphicx}
\fi
\usepackage{amssymb}
\usepackage{amsfonts}
\usepackage{xspace}
\usepackage{xcolor}
\usepackage{intmacros}
\begin{document}

%
\title{Sparsity Constrained Distributed Unmixing of Hyperspectral Data}
%
%
%

\author{Sara~Khoshsokhan,~\IEEEmembership{Student Member,~IEEE,}
        Roozbeh~Rajabi,~\IEEEmembership{Member,~IEEE,}
        and~Hadi~Zayyani,~\IEEEmembership{Member,~IEEE}
\thanks{The authors are with the Communications and Electronics Department, Faculty of Electrical and Computer Engineering, Qom University of Technology, Qom, Iran (e-mail: khoshsokhan.s@qut.ac.ir; rajabi@qut.ac.ir; zayyani@qut.ac.ir). Dr. Zayyani is also with the School of Electrical and Computer Engineering, Shiraz University, Shiraz, Iran.}}


%
%

\markboth{IEEE JSTARS}%
{Khoshsokhan \MakeLowercase{\textit{et al.}}: Sparsity Constrained Distributed Unmixing of	Hyperspectral Data}
%



\maketitle

\begin{abstract}
	Spectral unmixing (SU) is a technique to characterize mixed pixels in hyperspectral images measured by remote sensors. Most of the spectral unmixing	algorithms are developed using the linear mixing models. To estimate endmembers and fractional abundance matrices in a blind problem, nonnegative matrix factorization (NMF) and its developments are widely used in the SU problem. One of the constraints which was added to NMF is sparsity, that was regularized by $ L_ {q} $ norm. In this paper, a new algorithm based on distributed optimization is suggested for spectral unmixing. In the proposed algorithm, a network including single-node clusters is employed. Each pixel in the hyperspectral images is considered as a node in this network. The sparsity constrained distributed unmixing is optimized with diffusion least mean p-power (LMP) strategy, and then the update equations for fractional abundance and signature matrices are obtained. Afterwards the proposed algorithm is analyzed for different values of LMP power and $ L_ {q} $ norms. Simulation results based on defined performance metrics illustrate the advantage of the proposed algorithm in spectral unmixing of hyperspectral data compared with other methods.
\end{abstract}


%
\IEEEpeerreviewmaketitle

\section{Introduction}
%
%
%
%
\IEEEPARstart{R}{emote} sensors gather data by detecting the energy that is reflected from the Earth. Recent advances in remote sensing have paved the way for the development of hyperspectral sensors. One of the challenges in hyperspectral imaging is mixed pixels. Pixels in a scene constituting a single material are called pure pixels and other pixels containing more than one material are called mixed pixels \cite{Agg16}. So, the recorded spectrum of a single pixel is a mixture of some material spectra in the scene, named endmembers. The contribution of each endmember for a given pixel is weighted by its fractional abundance \cite{Miao07}. Decomposition of the mixed pixels is known as spectral unmixing (SU) problem \cite{Mei11}. Most of the spectral unmixing methods are based on linear mixing model (LMM), in which it is assumed that the recorded spectrum of a particular pixel, is linearly mixed by endmembers which exist in the pixel. In return, several researches have done by adopting nonlinear mixing model. In this model, considered pixel is generated from a nonlinear function of fractional abundances of endmembers \cite{Heylen14}, \cite{yang2018}. If the number of endmembers that are present in the scene and its signatures, are unknown, the SU problem becomes a blind source separation (BSS) problem \cite{Qian11}. 

Several SU methods have been proposed in different models. Unmixing methods can be categorized into three class of approaches: geometrical, statistical and sparse regression based methods \cite{STARS12_Overview}.  Pixel purity index (PPI) \cite{Chang06}, N-FINDR \cite{Winter99}, simplex volume maximization \cite{Chan11}, convex cone analysis (CCA) \cite{Ifarraguerri99}, successive projections algorithm (SPA) \cite{Araujo01}, principal component analysis (PCA) \cite{Smith85}, vertex component analysis (VCA) \cite{Nascimento05}, \cite{Lopez12} and robust minimum volume simplex analysis \cite{TGRS17_RMVSA} are some of convex geometric methods. They are based on the pure pixel assumption, where the simplex volume is considered as a criterion for detection of endmembers. Also, some of the methods such as independent component analysis (ICA) \cite{Bayliss98} use statistical models to solve the SU problem. Approaches like SUnSAL and C-SUnSAL \cite{WHISPERS10_SUNSAL} that work based on variable splitting and augmented Lagrangian, robust sprase unmixing (RSU) using $L_{2,1}$ \cite{TGRS17_RobustSparseL21} and double reweighted sparse regression and total variation (DRSU-TV) \cite{GRSL17_DoubleReweightedTV} are examples of sparse regression based approaches.

 Nonnegative matrix factorization (NMF) \cite{Paatero94}, \cite{Lee99} is another practical method of unmixing, which decomposes the data into two nonnegative matrices. Recently, this basic method was developed by adding some constraints, such as the minimum volume constrained NMF (MVC-NMF) method \cite{Miao07}, graph regularized NMF (GNMF) \cite{JIRS15_Rajabi_SparseGraph} and manifold regularized sparse NMF (GLNMF) \cite{lu13}. GLNMF is a two steps approach including sparse constraint and graph regularization. NMF with local smoothness constraint (NMF-LSC) \cite{Yang15} and total variation regularized reweighted sparse NMF (TV-RSNMF) \cite{he17tvl} are the methods in which the total variation regularizer is embedded into the reweighted sparse NMF. Also, some new NMF-based algorithms such as robust collaborative NMF \cite{li2016}, estimate the number of endmembers, in addition to spectral signatures and fractional abundances. One of the constraints that has been used to improve performance of NMF methods is sparsity constraint applied to NMF cost function using $L_q$ regulaizers \cite{Qian11}, or using another constraint such as smooth and sparse regularization proposed in \cite{salehani2017}. Since the number of endmembers present at each mixed pixel is normally scanty compared with the number of total endmembers, the problem becomes sparse \cite{TGRS11_SparseUnmixing}. So, the abundance matrix has many zero elements, and it has a large degree of sparsity. Using $ L_ {1/2} $ regularization in NMF, the authors in \cite{Qian11} proposed the $ L_ {1/2} $-NMF algorithm, which enforces the sparsity of endmember abundances. Another method is proposed in \cite{IJRS18_l1l2_TV} based on difference of $L_1$ and $L_2$ norm in conjunction with total variation regualrization. Overall, NMF approaches that exploit two-block alternating optimization (AO) offer good performance to solve hyperspectral unmixing problem \cite{SPM14_SignalProcessingPerspective}. Recently, multilayer NMF (MLNMF) \cite{WHISPERS14_MLNMF, Rajabi15} and deep NMF \cite{TGRS18_DeepNMF} has been proposed to use in hyperspectral unmixing problems that can effectively decompose observation matrix in different layers to reduce reconstruction error.

As another approach, the distributed strategy has been used for utilization of neighborhood information as spatial information. This can be advantageous, because the neighboring pixels may be correlated. Spatial information has been used in different ways in spectral unmixing \cite{zhang18}, including total variation spatial regularization for sparse hyperspectral unmixing \cite{ior12}. There are some distributed strategies such as consensus strategies \cite{Tsitsiklis84}, incremental strategies \cite{Bertsekas97} and diffusion strategies \cite{TSP10_DiffusionLMS}. In this paper, a diffusion strategy is used because it has strong stability over adaptive networks \cite{Chen14,isprs17_distributedUnmixing}. Diffusion least mean p-power (LMP) strategy has been proposed in \cite{Wen13} and used in different applications like system identification \cite{SPL14_LMP}, robust sparse recovery \cite{TSP17_LMP}.

To solve a distributed problem, a network is considered. There are three types of networks: 1) a single-task network, that nodes estimate a common unknown and optimum vector, 2) a multitask network, in which each node estimates its own optimum vector and 3) a clustered multitask network includes clusters that each of them has to estimate a common optimum vector \cite{Chen14}. Unmixing problem is a multitask problem where each pixel is considered to be a node. Here, we first consider the general case, where there is a clustered multitask network and each cluster has an optimum vector (fractional abundance vector) that should be estimated. Then we will reduce that to a multitask case.

The main contributions of this paper are as follows,

\textbullet\ {Sparsity constraint has been added to the distributed method for spectral unmixing and LMP strategy is used,}

\textbullet\ {The cost function has been generalized using $q_1$ and $q_2$ norms,}

\textbullet\ {Simultaneous estimation of spectral signatures and fractional abundances using NMF method has been added to distributed unmixing.}

This paper is organized as follows. In section II, we introduce the proposed method and optimize it. Section III includes an introduction of datasets. Section IV provides simulation results and the last section gives conclusions.

\section{Distributed Unmixing of Hyperspectral Data With Sparsity Constraint}
In this section, a new method that utilizes sparsity constraint and neighborhood information is proposed. First, we express linear mixing model in subsection II.A, then the sparsity constrained distributed cost functions are formulated in II.B, and finally, we use them to solve SU problem in II.C.
\subsection{Linear Mixing Model (LMM)}
To solve the SU problem, we focus on a simple but representative model, named LMM. In this model, there exists a linear relation between the endmembers that are weighted by their fractional abundances, in the scene. Mathematically, this model is described as $\mathbf{y}_k=\mathbf{A}\mathbf{s}_k+\mathbf{v}_k$ and the matrix form of this equation can be written as:
\begin{equation}
\label{modelmatrix}
\mathbf{Y}=\mathbf{A}\mathbf{S}+\mathbf{V}
\end{equation}
where $\mathbf{y}_k$ is an $L\times1$ observed data vector for $k$th pixel, $\mathbf{Y}=[\mathbf{y}_1,\mathbf{y}_2,..,\mathbf{y}_N]$ is an $L\times N$ observed data matrix, $\mathbf{A}$ is the $L\times c$ signature matrix, $\mathbf{s}_k=[s_k(1),s_k(2),..,s_k(c)]^T$ is the $c\times1$ fractional abundance vector for $k$th pixel, $\mathbf{S}=[\mathbf{s}_1,\mathbf{s}_2,..,\mathbf{s}_N]$ is the $c\times N$ fractional abundance matrix, $\mathbf{v}_k$ is assumed as an $L\times1$ additive noise vector of $k$-th pixel of the image and $\mathbf{V}=[\mathbf{v}_1,\mathbf{v}_2,..,\mathbf{v}_N]$ is an $L\times N$ additive noise matrix where $c$, $L$ and $N$ denote the number of endmembers, bands and pixels, respectively.

In the SU problem, fractional abundance vectors have two constraints in each pixel, abundance sum to one constraint (ASC) and abundance nonnegativity constraint (ANC) \cite{Ma14}, which are as follows, for $c$ endmembers in a scene. 
\begin{equation}
\sum\limits_{n=1}^c \mathbf{s}_{k}(n)=1;\;\;\mathbf{s}_{k}(n)\geq0,n=1,...,c
\end{equation}
Where $\mathbf{s}_{k} (n)$ is the fractional abundance of the $n$-th endmember in the $k$-th pixel of the image. Fractional abundances of endmembers are nonnegative values and endmembers present in a mixed pixel cover all area of that mixed pixel, hence they add up to one. Note that, in a BSS problem, only the observed vector is known and we aim to determine the two other matrices ($\mathbf{A}$ and $\mathbf{S}$ in equation (\ref{modelmatrix})).
\subsection{Distributed Cost Functions and Optimization}
As explained earlier, three types of networks containing single task, multitask and clustered multitask networks are supposed. First, $N$ nodes are considered in a clustered multitask network and an optimum vector at node $k$ is estimated. A global cost function using LMP with $p$ power \cite{Wen13}, $J^{global}(\mathbf{s}_k(n))$, is defined as follows:
\begin{equation}
\label{eq: 4}
J^{global} (\mathbf{s}_1,\mathbf{s}_2,...,\mathbf{s}_N)=\sum\limits_{k=1}^N \mathbb{E}\{|\mathbf{y}_k -\mathbf{A} \mathbf{s}_k|^{p}\}
\end{equation}
where $\mathbb{E}$ is the expectation operator. Then, to minimize the cost function, the following equation is written, using the iterative steepest-descent solution \cite{Cattivelii10}:
\begin{equation}
\label{eq: 5}
\mathbf{s}_k^i=\mathbf{s}_k^{i-1}-\mu \Big(\bigtriangledown_{\mathbf{s}} J(\mathbf{s}_k^{i-1})\Big)^*
\end{equation}
where $\mathbf{s}_k^{i}$ is the fractional abundance vector of the $k$th node in the $i$th iteration, $\mu>0$ is a step-size parameter, and the algorithm make small jumps, using an optimum value of $\mu$. This optimum value causes stability and depends on the cost function. The algorithm will diverge with a too large value of $\mu$, and will take a long time to converge with a too small value. $i$ is iteration number and $\bigtriangledown_{\mathbf{s}}$ is the gradient operator with respect to $\mathbf{s}$, $*$ is the complex conjugate operator, then after computing complex gradient and substituting it into (\ref{eq: 5}), the following iterative equation is obtained:
\begin{equation}
\label{eq: 6}
\mathbf{s}_k^{i}=\mathbf{s}_k^{i-1}+\mu \sum\limits_{k=1}^N \Big(\mathbf{A}^T |\mathbf{e}_k|^{p-2} \mathbf{e}_k\Big)
\end{equation}
where $\mathbf{e}_k=\mathbf{y}_k-\mathbf{A} \mathbf{s}_k$ is the error signal. 

Since the neighborhood information has not been used yet, the equation (\ref{eq: 6}) is not distributed. In a distributed network, relationships between neighboring nodes are used to improve accuracy. In this article, we utilize the $L_{q_1}$ norm:
\begin{equation}
\label{eq: 8}
\Delta(\mathbf{s}_k,\mathbf{s}_l)=||\mathbf{s}_k-\mathbf{s}_l||_{q_1}
\end{equation}

There are different criteria to measure sparsity of hyperspectral images \cite{jstars14_sparseGPU,Qian11}. In this paper we generalized sparsity norm and used the $L_{q_2}$ regularizer for sparsity constraint:
\begin{equation}
\label{eq: 9}
||\mathbf{s}_k||_{q_2}=\big( \sum\limits_{n=1}^c \mathbf{s}_{k}^{q_2} (n)\big)^{1/q_2}
\end{equation}

Note that, the solution determined from global cost function, needs to have access to information over the entire network, but the nodes can be considered to have availability only to information of their neighbors. Thus, for solving this problem, the following local cost function for $k$th pixel is defined, using LMP and adding the (\ref{eq: 8}) and (\ref{eq: 9}) constraints:
\begin{equation}
\begin{aligned}
\label{eq: 10}
&J^{local} (\mathbf{s}_k)=\sum\limits_{m\in \mathcal{N}_k \cap C(k)}  \mathbb{E}\{|y_m(l)-\mathbf{a}_{l} \mathbf{s}_k|^{p}\}&\\
&+\eta \sum\limits_{j\in \mathcal{N}_k \backslash C(k)} \rho_{kj}  ||\mathbf{s}_{k}-\mathbf{s}_{l}||_{q_1} + \lambda ||\mathbf{s}_k||_{q_2}&
\end{aligned}
\end{equation}

\begin{figure}[!t]
	\centering
	\includegraphics[draft=false,width=2in]{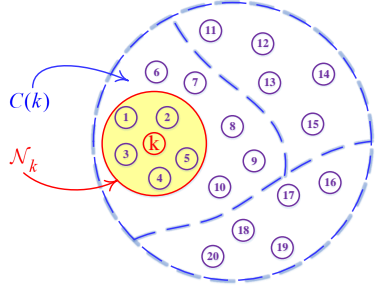}
	\caption{Illustration of neighborhood and clusters in a typical clustered multitask network.}
	\label{fig:neighborcluster}
\end{figure}
where $\mathbf{a}_{l}$ is a $1\times c$ vector equals to $A(l,:)$, the symbol $\backslash$ is the set difference, $\mathcal{N}_k$ shows nodes that are in the neighborhood of node $k$, that is in the cluster $C(k)$ (see \figurename{~\ref{fig:neighborcluster}}). $\eta>0$  denotes a regularization parameter \cite{Chen14}, that controls the effect of neighborhood term, $\lambda$ is a scalar value that weights the sparsity function \cite{Qian11},  and the nonnegative coefficients $\rho_{kj}$ are normalized spectral similarity which is obtained from correlation of data vectors \cite{Chen14}. The coefficients are computed as introduced in \cite{Qian11}, \cite{Chen14}:

\begin{equation}
\label{eq: lambda}
\lambda = \frac{1}{\sqrt{L}} \sum\limits_{l=1}^{L}  \frac{\sqrt{N}-||\mathbf{x}_l||_1/||\mathbf{x}_l||_2}{\sqrt{N-1}}
\end{equation}
where $\mathbf{x}_l$ is a $N\times1$ vector denoting the $l$th band of the hyperspectral image ($\mathbf{x}_l=(\mathbf{Y}(l,:))^T$).
\begin{equation}
\label{eq: rho}
\rho_{kj} = \frac{\theta (\mathbf{y}_k,\mathbf{y}_j)}{\sum\limits_{l \in\mathcal{N}_k^-} \theta (\mathbf{y}_k,\mathbf{y}_l)}
\end{equation}
where $\mathcal{N}_k^-$ includes neighbors of node $k$ except itself, and $\theta$ is computed as \cite{Chen14}:
\begin{equation}
\label{eq: theta}
\theta (\mathbf{y}_k,\mathbf{y}_j) = \frac{\mathbf{y}_k^T \mathbf{y}_j}{||\mathbf{y}_k|| ||\mathbf{y}_j||}
\end{equation}

Now, to minimize the cost function of (\ref{eq: 10}), using steepest-descent in equation (\ref{eq: 5}), we have:
\begin{equation}
\begin{aligned}
\label{eq: 19}
&\mathbf{s}_k^{i+1} = \mathbf{s}_k^{i}+ \mu \sum\limits_{l\in \mathcal{N}_k \cap C(k)} \big(\mathbf{A}^T |\mathbf{e}_k|^{p-2} \mathbf{e}_k\big)&\\
&+ \mu \eta \sum\limits_{j\in \mathcal{N}_k\backslash C(k)} \rho_{kj}  \frac{\big( \mathbf{s}_k^{i}- \mathbf{s}_j^{i}\big) |\mathbf{s}_k^{i}- \mathbf{s}_j^{i}|^{q_1-2}}{||\mathbf{s}_k^{i}- \mathbf{s}_j^{i}||_{q_1}^{q_1-1}}- \mu \lambda \frac{\big( \mathbf{s}_k^{i}\big) |\mathbf{s}_k^{i}|^{q_2-2}}{||\mathbf{s}_k^{i}||_{q_2}^{q_2-1}} &
\end{aligned}
\end{equation}

As explained earlier, the SU is a multitask problem that is adopted as \figurename{~\ref{fig:neighborcluster}} using single node clusters. Therefore, (\ref{eq: 19}) with adoption of LMP strategy, is simplified to:
\begin{equation}
\begin{aligned}
\label{eq: s}
&\mathbf{s}_k^{i+1}= \mathbf{s}_k^{i}+ \mu  \big(\mathbf{A}^T|\mathbf{e}_k|^{p-2} \mathbf{e}_k\big)&\\
&+\mu \eta \sum\limits_{j\in \mathcal{N}_k^-} \rho_{kj} \frac{\big( \mathbf{s}_k^{i}- \mathbf{s}_j^{i}\big) |\mathbf{s}_k^{i}- \mathbf{s}_j^{i}|^{q_1-2}}{||\mathbf{s}_k^{i}- \mathbf{s}_j^{i}||_{q_1}^{q_1-1}}-\mu \lambda \frac{\big( \mathbf{s}_k^{i}\big) |\mathbf{s}_k^{i}|^{q_2-2}}{||\mathbf{s}_k^{i} ||_{q_2}^{q_2-1}}&
	\end{aligned}
	\end{equation}
	Hence, this recursive equation can be used to update fractional abundance vectors in SU problems.
\subsection{Spectral Unmixing Updating Equations}
Similar to the NMF algorithm, the least mean p-power error should be minimized with respect to the signatures and abundances matrices, subject to the non-negativity constraint \cite{Lee01}. So, the following equation is denoted, using matrix notation:
\begin{equation}
\label{eq: 21}
\min\limits_{\mathbf{S},\mathbf{A}>0} ||\mathbf{Y}-\mathbf{A}\mathbf{S}||_F^p
\end{equation}
where $\mathbf{A}$ and $\mathbf{S}$ are the $L\times c$ signature and $c\times N$ fractional abundances matrices, respectively, and Y denotes the $L\times N$ Hyperspectral data matrix. Then, based on described equations of the sparsity constrained distributed unmixing, the neighborhood and sparsity terms are added to (\ref{eq: 21}) as follows:
\begin{equation}
\label{eq: 22}
||\mathbf{Y}-\mathbf{A}\mathbf{S}||_F^p + \eta \sum \limits_{k=1}^N \sum\limits_{j\in \mathcal{N}_k} \rho_{kj} ||\mathbf{s}_k - \mathbf{s}_j||_{q_1}+\lambda \sum \limits_{k=1}^N ||\mathbf{s}_{k}||_{q_2}
\end{equation}
This cost function is minimized with respect to $\mathbf{A}$, using multiplicative update rules \cite{Lee01}, then recursive equation of signature matrix is obtained as:
\begin{equation}
\label{eq: A}
\mathbf{A}^{i+1}=\mathbf{A}^i*\frac{\mathbf{Y}\mathbf{S}^T} {\mathbf{A}\mathbf{S}\mathbf{S}^T} 
\end{equation}
And the recursive equation of fractional abundances has been obtained already in accordance with (\ref{eq: s}) as follows:
\begin{equation}
\begin{aligned}
\label{eq: s1}
&\mathbf{s}_k^{i+1}= P^+\Big[\mathbf{s}_k^{i}+ \mu  \big(\mathbf{A}^T|\mathbf{e}_k|^{p-2} \mathbf{e}_k\big)&\\
&+\mu \eta \sum\limits_{l\in \mathcal{N}_k^-} \rho_{kl} \frac{\big( \mathbf{s}_k^{i}- \mathbf{s}_l^{i}\big) |\mathbf{s}_k^{i}- \mathbf{s}_l^{i}|^{q_1-2}}{||\mathbf{s}_k^{i}- \mathbf{s}_l^{i}||_{q_1}^{q_1-1}}-\mu \lambda \frac{\big( \mathbf{s}_k^{i}\big) |\mathbf{s}_k^{i}|^{q_2-2}}{||\mathbf{s}_k^{i}||_{q_2}^{q_2-1}}\Big]&
\end{aligned}
\end{equation}
where $P^+$ operator projects vectors onto a simplex, that adopt the ASC and ANC constraints for abundance vectors. The $P^+$ operator has been defined in \cite{Chen11}. Another significant point in implementation of the algorithm is stopping criteria. This approach will be stopped until the maximum number of iteration ($T$), or the following stopping criteria is reached.
\begin{equation}
\label{eq: stop}
||J_{new}-J_{old}||<\epsilon
\end{equation}
where $J_{new}$ and $J_{old}$ are cost function values for two consecutive iterations and $\epsilon$ has been set to $10^{-8}$ in our experiments. Now, the proposed approach is summarized in Algorithm 1.
\begin{algorithm}
	\SetKwInOut{Input}{input}
	\SetKwInOut{Output}{output}
	\SetKwInput{Initialization}{Initialisation}
	\Input{Hyperspectral data matrix ($\mathbf{Y}$)\\
	Parameters: $N$,$c$,$L$,$\mu$,$\eta$,$p$,$q_1$ and $q_2$}
	\Output{Estimated fractional abundance and signature matrices ($\mathbf{S}$ and $\mathbf{A}$)}
	\Initialization{Initialize the $\mathbf{A}$ and $\mathbf{S}$ matrices by random matrices or the outcome of VCA-FCLS algorithm. Compute $\lambda$ and $\rho$ values from (\ref{eq: lambda}) and (\ref{eq: rho}).}
	\While{the maximum number of iteration ($T$) or stopping criteria in (\ref{eq: stop}) has been reached}
	{
		a. Update $\mathbf{A}$, using  (\ref{eq: A})\;
		b. Update $\mathbf{s}_k$ for all pixels, by applying  (\ref{eq: s})\;
		c. Adopt $P^+$ operator for ASC and ANC constraints\;
   end}
	\caption{Sparsity constrained distributed unmixing}
	\label{algorithm}
\end{algorithm}

A usual way for evaluation of algorithms is computational cost. \tablename{~\ref{t}} shows comparison of computational complexity order for each iteration between NMF, $L_{1/2}$-NMF, GLNMF, TV-RSNMF, distributed unmixing and the proposed method. The complexity of sparsity constrained distributed unmixing with $p=q_1=q_2=2$ is presented in the last row. It should be noticed, the proposed approach becomes more complicated with non integer values of $p$, $q_1$ and $q_2$.

\begin{table*}
		\centering
		\caption{
			Comparison between computational complexity of different methods.
		}
		\scalebox{0.9}{
		\begin{tabular}{|p{1.7cm}|p{4.15cm}|p{7cm}|p{1.25cm}|p{2.25cm}|}
			\hline
			 Methods & Addition & Multiplication & Division & Complexity order\\
			\hline \hline
			NMF & $2NcL+2c^2(N+L)$ & $2NcL+2c^2(N+L)+c(N+L)$ & $c(N+L)$ & $\mathcal{O}(NcL)$\\
			\hline
			$L_{1/2}$-NMF & $2NcL+2c^2(N+L)$ & $2NcL+2c^2(N+L)+c(N+L)+(cN)^2$ & $c(N+L)$ &$\mathcal{O}(NcL+(cN)^2)$\\
			\hline
			GLNMF & $2NcL+2c^2(N+L)$ $+Nc(k+4)$ & $2NcL+2c^2(N+L)+c(kN+2N+L)+(cN)^2$ & $c(N+L)$ &$\mathcal{O}(NcL+(cN)^2)$\\
			\hline
			TV-RSNMF & $2NcL+2c^2(N+L)$ & $2NcL+2c^2(N+L)+c(N+L)$ & $c(2N+L)$ & $\mathcal{O}(NcL)$\\
			\hline
			Distributed & $NcL+c^2(N+L)+N(c+L+k)$ & $3NcL+c^2(N+L)+cL$ & $cL$ & $\mathcal{O}(NcL)$\\
			\hline
			Proposed alg. & $NcL+c^2(N+L)+N(4c+2L+k)$ & $4NcL+c^2(N+L)+c(N+L)+Nc(q_1+q_2-4)+NL(p-2)$ & $cL$ & $\mathcal{O}(NcL)$\\
			\hline
		\end{tabular}}
		\label{t}
\end{table*}

\section{Datasets}
In this paper, the proposed algorithm is tested on synthetic and real datasets. This section introduces a real dataset that recorded with hyperspectral sensors and a synthetic dataset that are generated using spectral libraries.
\subsection{Synthetic Images}
To generate synthetic data, some spectral signatures are chosen from a digital spectral library (USGS) \cite{Clark07}, that include 224 spectral bands, with wavelengths from 0.38$\mu$$m$ to 2.5$\mu$$m$. Size of intended images is 64$\times$64, and one endmember is contributed in spectral signature of each pixel, randomly. Pixels of this image are pure, so to have an image containing mixed pixels, a low pass filter is considered. It averages from abundances of endmembers in its window, so that the LMM would be confirmed. The size of this window controls degree of mixing \cite{Miao07}. With smaller dimension of the window and more endmembers in the image, degree of sparsity is increased.
\subsection{Real Data}
\subsubsection{AVIRIS Cuprite}
\begin{figure}
	\centering
	\subfigure[]{
		\includegraphics[draft=false,height=3cm]{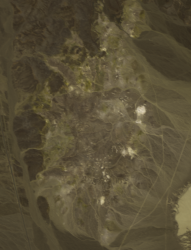}
		\label{fig:rgbAv}
	}
	\subfigure[]{
		\includegraphics[draft=false,height=3cm]{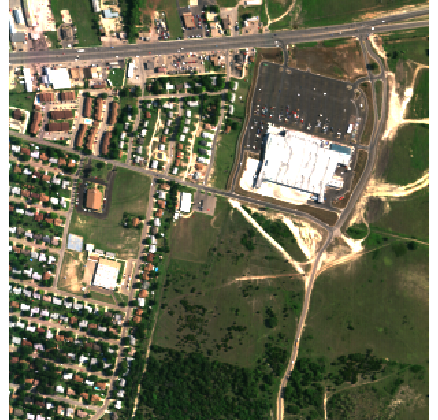}
		\label{fig:rgbHy}
	}
	\caption{(a) Pseudo color image of AVIRIS Cuprite data scene. The bands used as RGB channel are bands (40,20,10) of original 224 bands image and (b) pseudo color image of HYDICE Urban data scene. The bands used as RGB channel are bands (49,35,18) of original 210 bands image.}
	\label{fig:realData}
\end{figure}

The real dataset that the proposed method is applied on it, is hyperspectral data captured by the Airborne Visible/Infrared Imaging Spectrometer (AVIRIS) over Cuprite, Nevada. This dataset has been used since the 1980s. AVIRIS spectrometer has 224 channels and covers wavelengths from 0.4$\mu$$m$ to 2.5$\mu$$m$. Its spectral and spatial resolutions is about 10$nm$ and 20$m$, respectively \cite{Green98}. 188 bands of these 224 bands are used in the experiments. The other bands (covering bands 1, 2, 104-113, 148-167, and 221-224) have been removed which are related to water-vapor absorption or low SNR bands. \figurename{~\ref{fig:rgbAv}} illustrates a pseudo color image of this dataset.
\subsubsection{HYDICE Urban}
Now, turn our attention to another real dataset that is urban HYDICE hyperspectral image. There are 210 bands in this dataset, that covers wavelengths from 0.4$\mu$$m$ to 2.5$\mu$$m$. After removing water-vapor absorption or low SNR bands (including 1-4, 76, 87, 101-111, 136-153, and 198-210), only 162 bands are used in the experiments. There are just 4 distinguished materials in HYDICE Urban image, asphalt, roof, tree and grass \cite{He16}. \figurename{~\ref{fig:rgbHy}} illustrates a pseudo color image of this real dataset.

\section{Experiments and Results}
In this section firstly evaluation criteria are introduced and then the proposed algorithm is justified and parameter selection is done using experiments on synthetic data. Then to verify results of the proposed algorithm on real datasets, it is applied on AVIRIS cuprite and HYDICE urban datasets.
\subsection{Evaluation Criteria}
In this section, for quantitative comparison between the proposed method with different values of $p$, $q_1$ and $q_2$, also between proposed and the other methods, the performance metrics such as spectral angle distance ($SAD$) and abundance angle distance ($AAD$) \cite{Miao07} are used. They are defined as:
\begin{equation}
\label{eq: 25}
SAD= \cos^{-1} \Big(\frac{\mathbf{a}^T \hat{\mathbf{a}}}{||\mathbf{a}|| ||\hat{\mathbf{a}}||}\Big);\; AAD= \cos^{-1} \Big(\frac{\mathbf{s}^T \hat{\mathbf{s}}}{||\mathbf{s}|| ||\hat{\mathbf{s}}||}\Big)
\end{equation}
where $\hat{\mathbf{a}}$ is the estimation of spectral signature vectors and $\hat{\mathbf{s}}$ is the estimation of fractional abundance vectors.
\begin{figure}[!t]
	\centering
	\includegraphics[draft=false,width=1.5in]{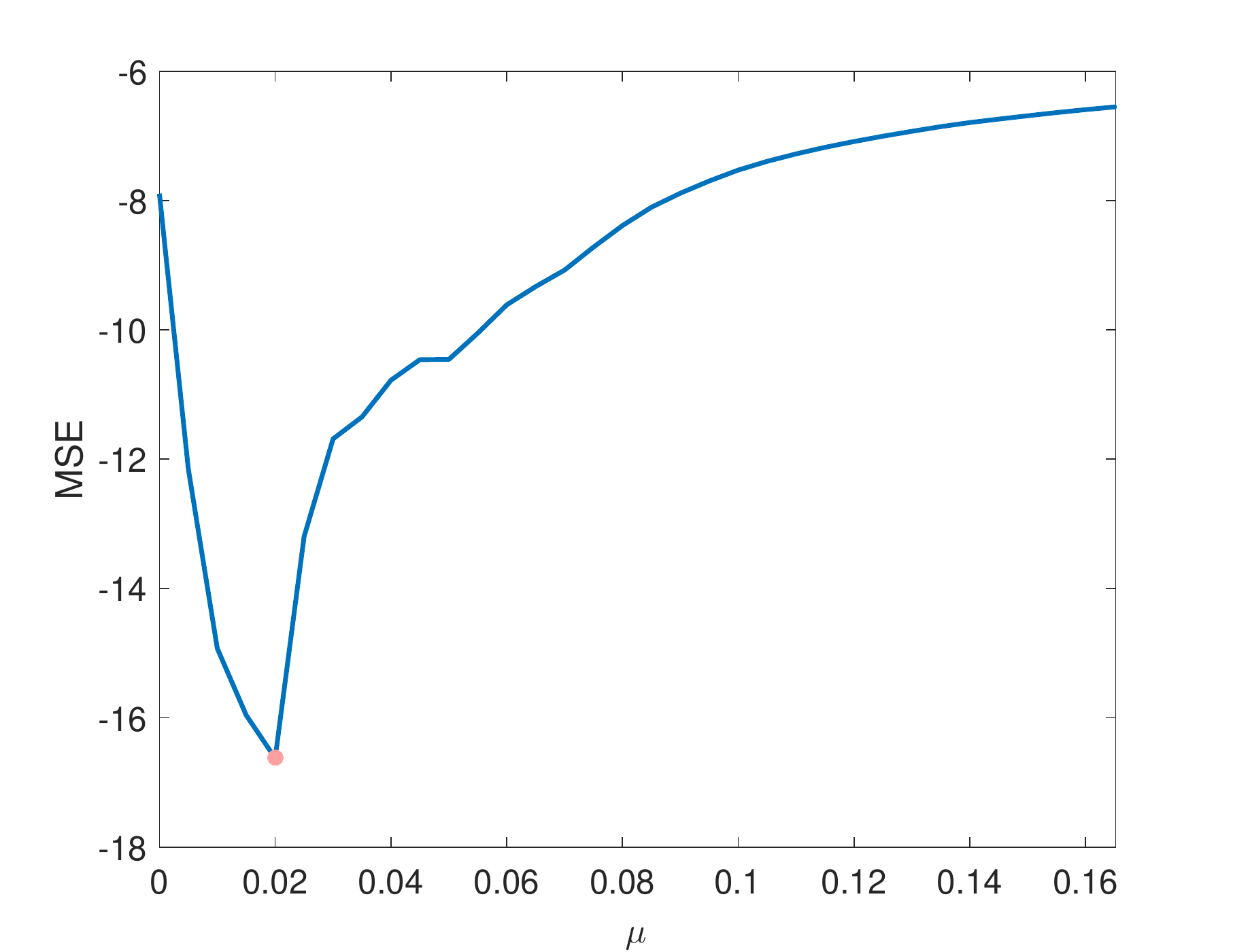}
	\caption{Logarithm of MSE versus $\mu$ in the obtained range, for selecting the best value of $\mu$, on synthetic dataset. The red point shows minimum of MSE value in $\mu=0.02$.}
	\label{mu}
\end{figure}
\begin{figure}[!ht]
	\centering
	\subfigure[]{
		\includegraphics[draft=false,width=3.7cm]{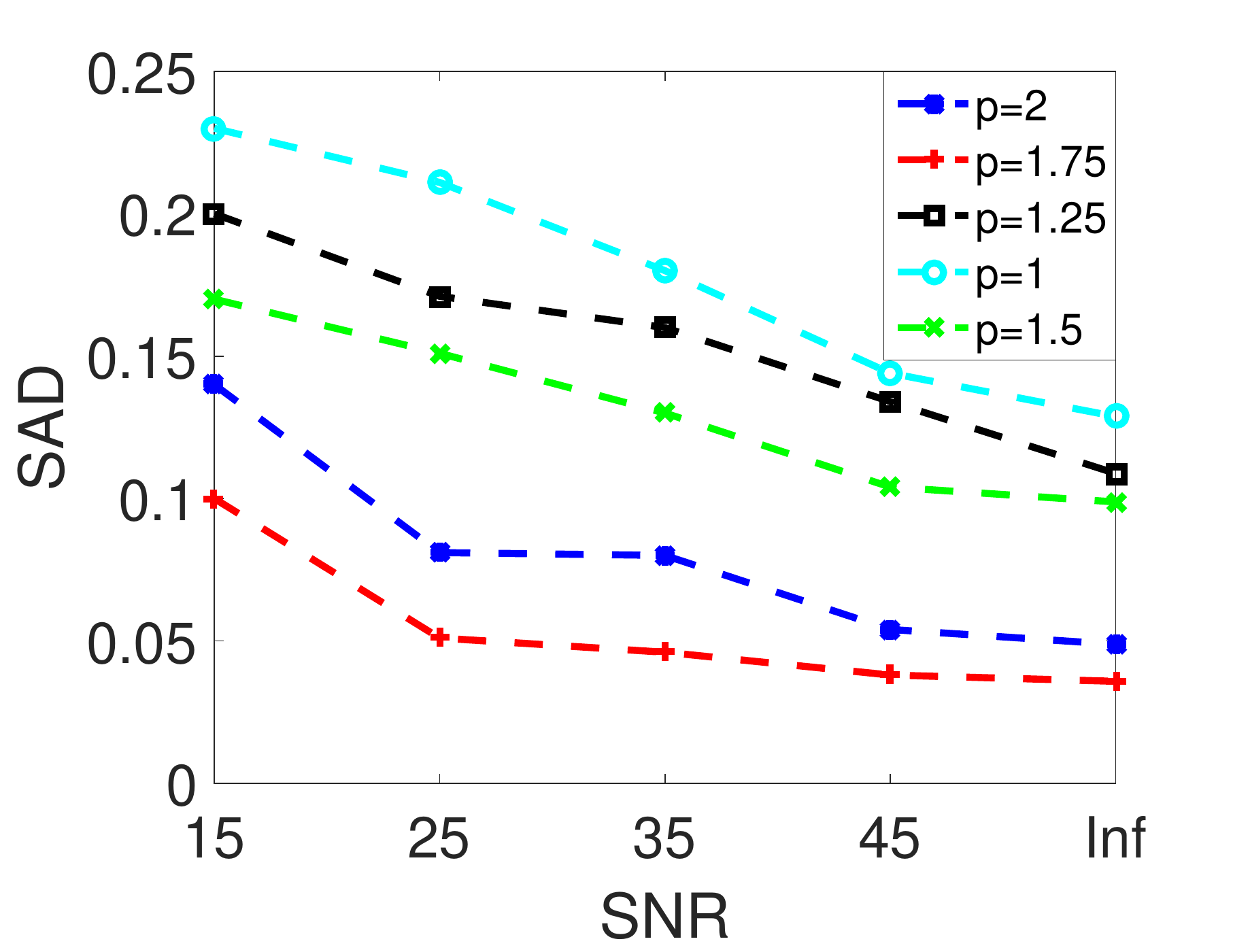}
		\label{fig:sub1}
	}
	\subfigure[]{
		\includegraphics[draft=false,width=3.7cm]{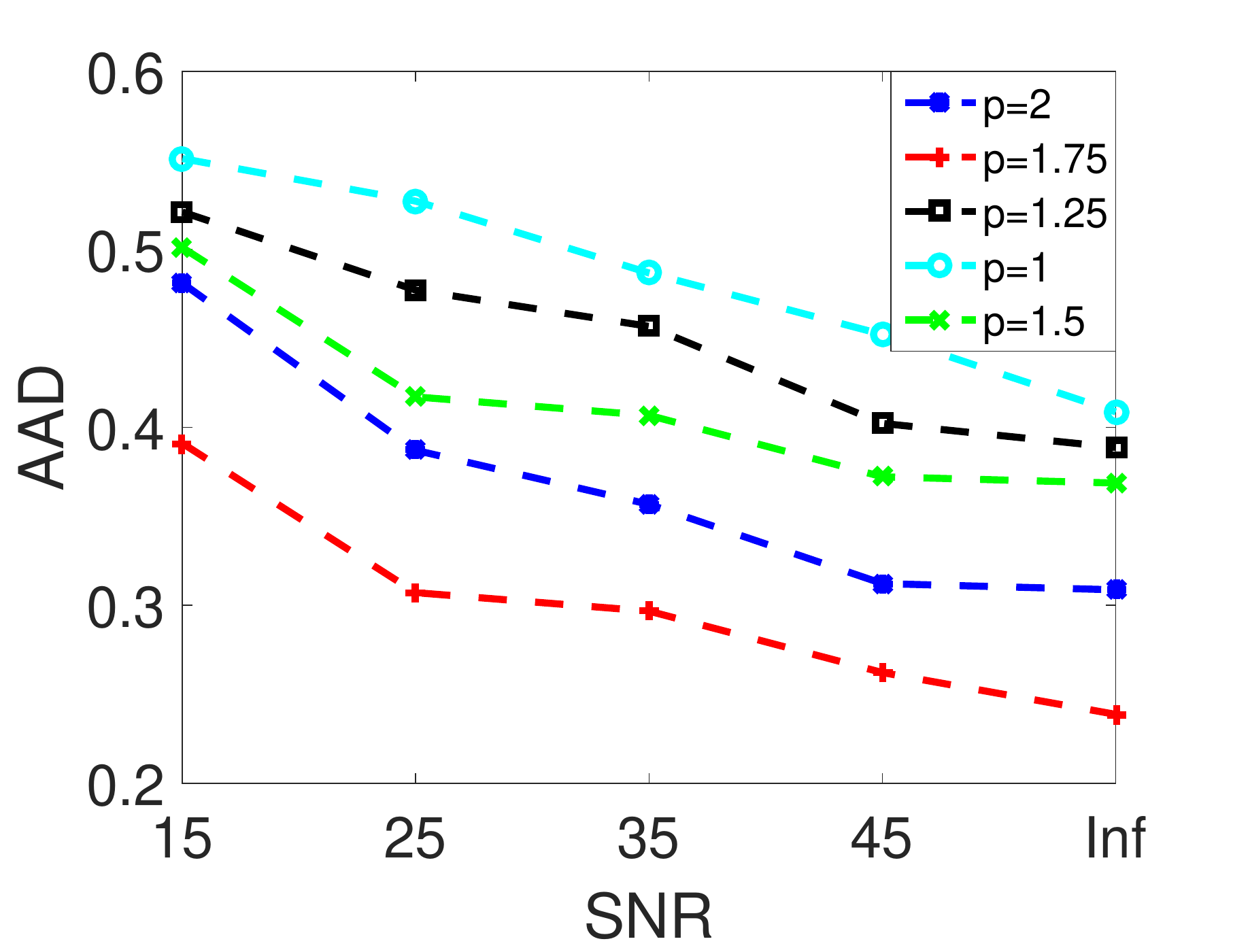}
		\label{fig:sub2}
	}
	
	\subfigure[]{
		\includegraphics[draft=false,width=3.7cm]{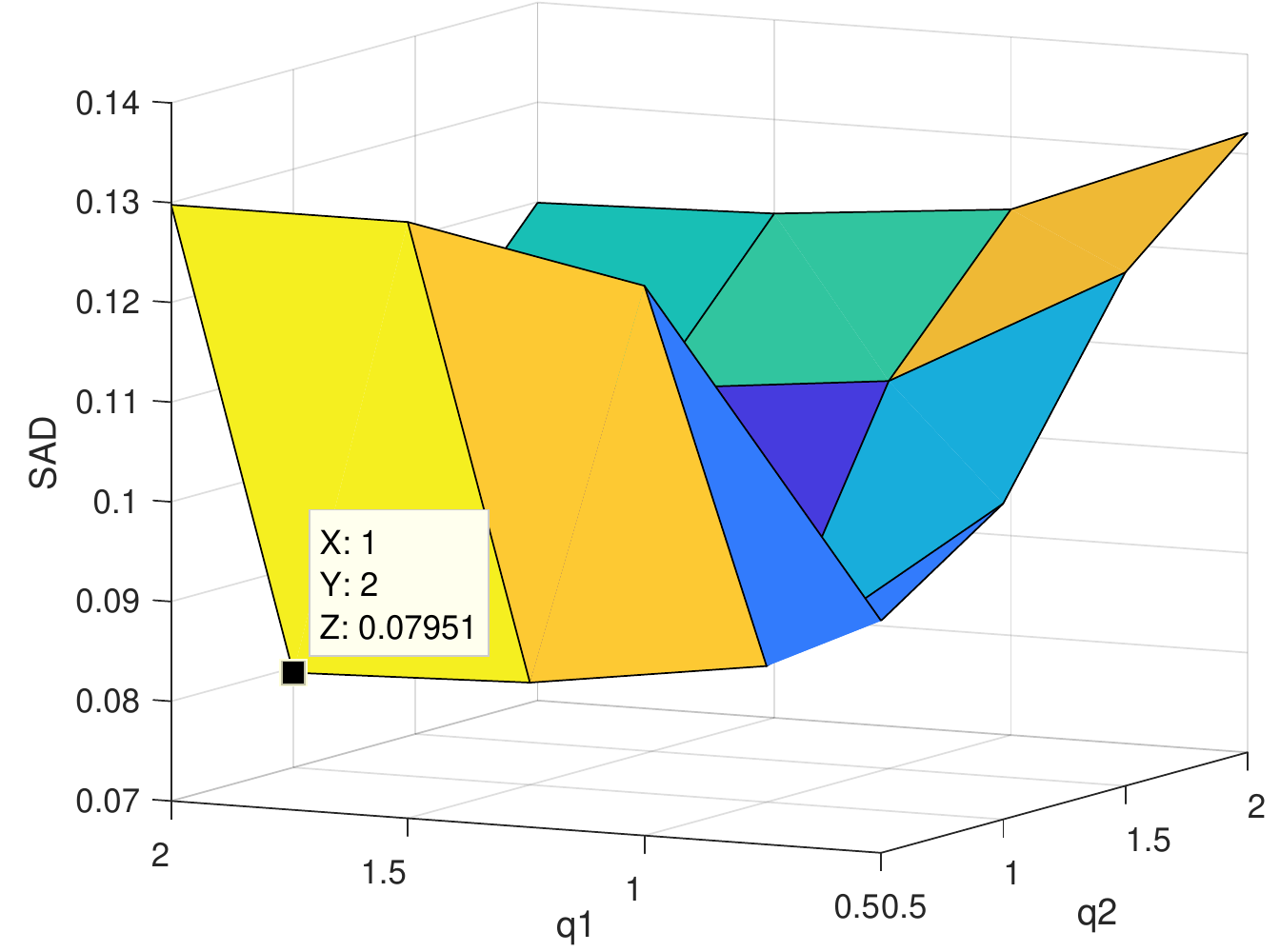}
		\label{fig:sub3}
	}
	\subfigure[]{
		\includegraphics[draft=false,width=3.7cm]{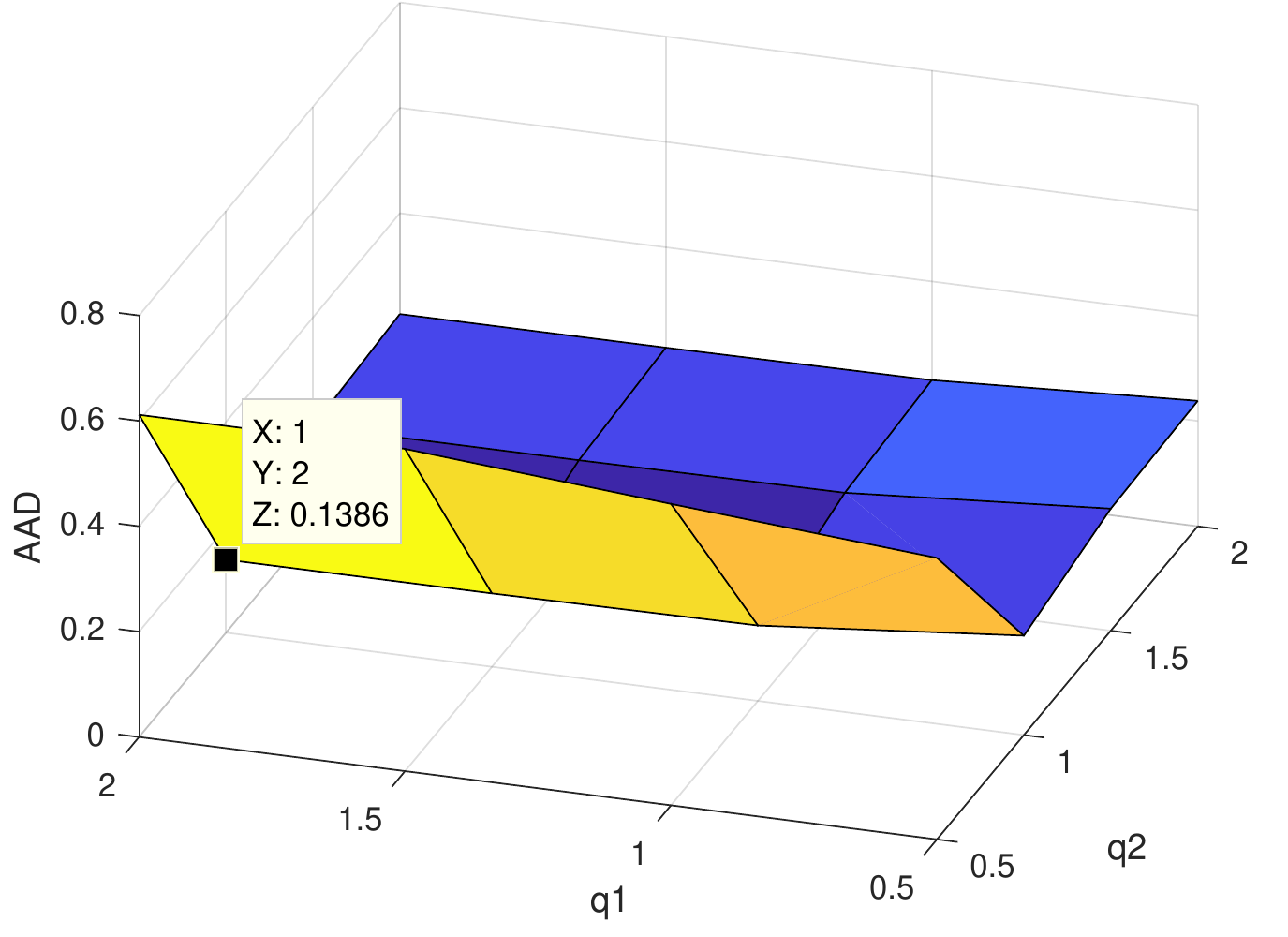}
		\label{fig:sub4}
	}
	\subfigure[]{
		\includegraphics[draft=false,width=3.7cm]{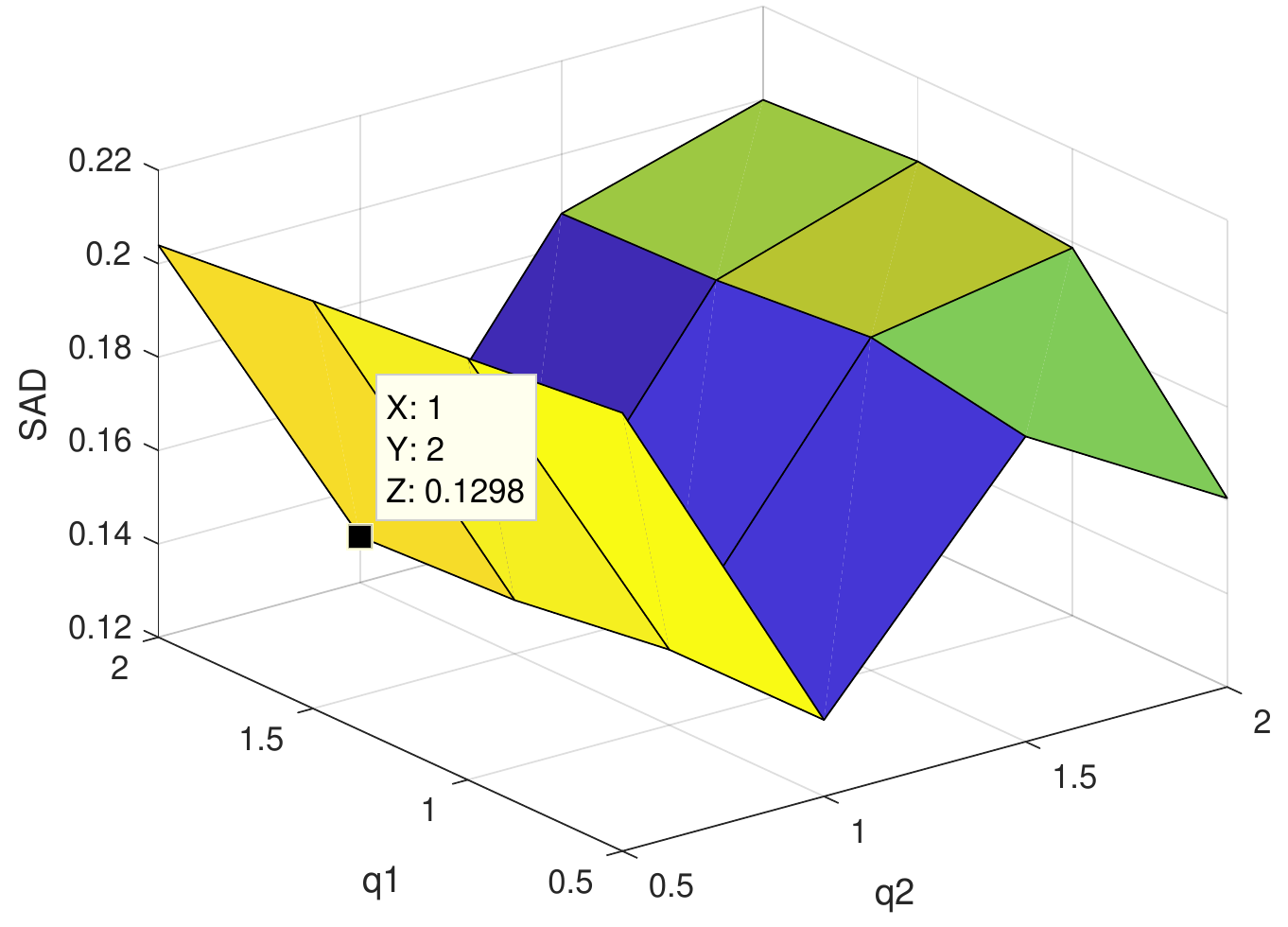}
		\label{fig:sub5}
	}
	\subfigure[]{
		\includegraphics[draft=false,width=3.7cm]{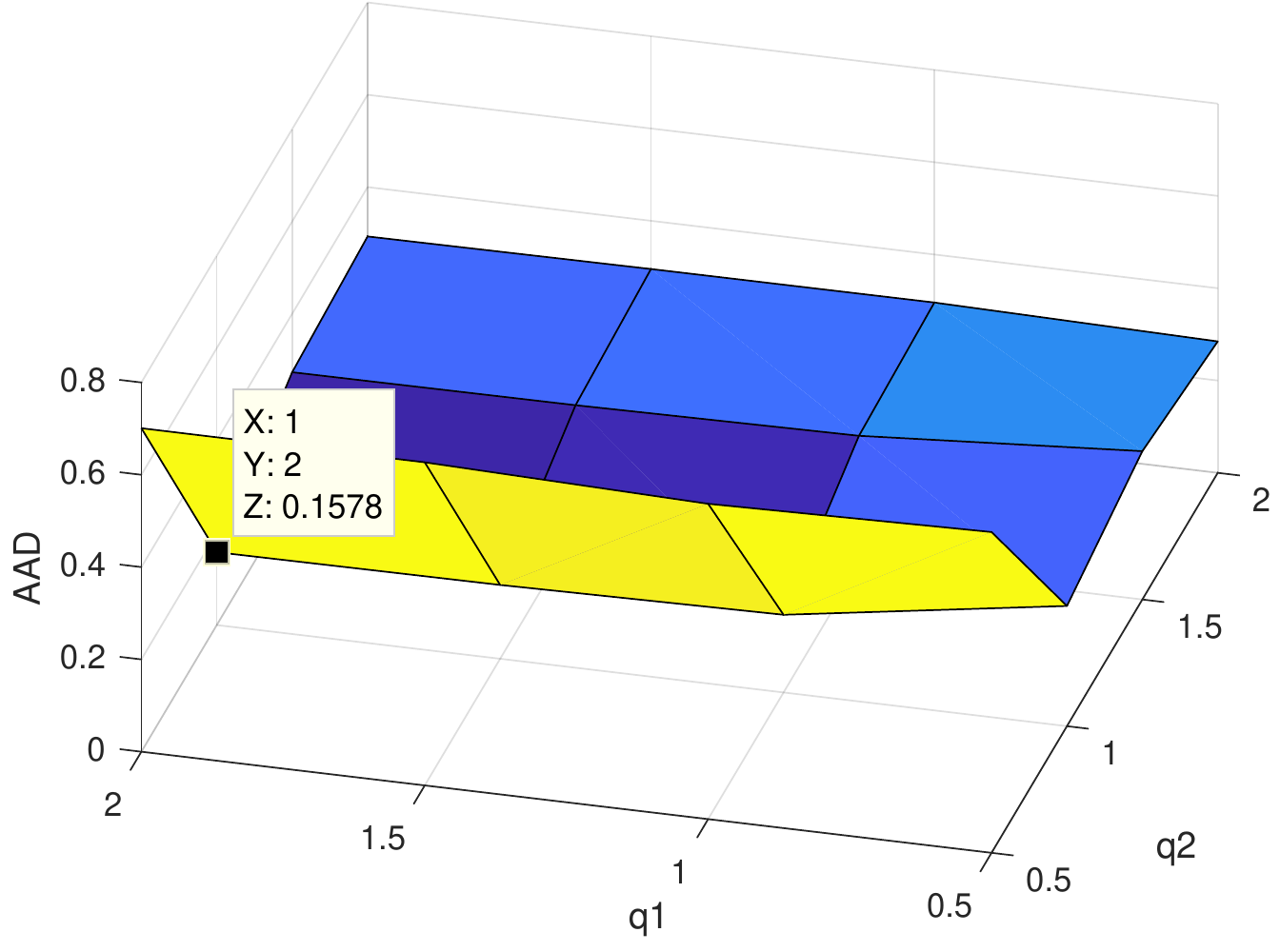}
		\label{fig:sub6}
	}
	\caption{(a) The $SAD$ and (b) $AAD$ performance metrics of proposed algorithm with different $p$ values versus SNR, with the same values of $q_1=2$ and $q_2=2$, (c) The $SAD$ and (d) $AAD$ performance metric of proposed algorithm with different $q_1$ and $q_2$ values when $p=1.75$, (e) The $SAD$ and (f) $AAD$ performance metric of proposed algorithm with different $q_1$ and $q_2$ values when $p=2$, using VCA initialization and applied on synthetic data.}
	\label{pq}
\end{figure}
\begin{figure}
	\centering
	\subfigure[]{
		\includegraphics[draft=false,width=4cm]{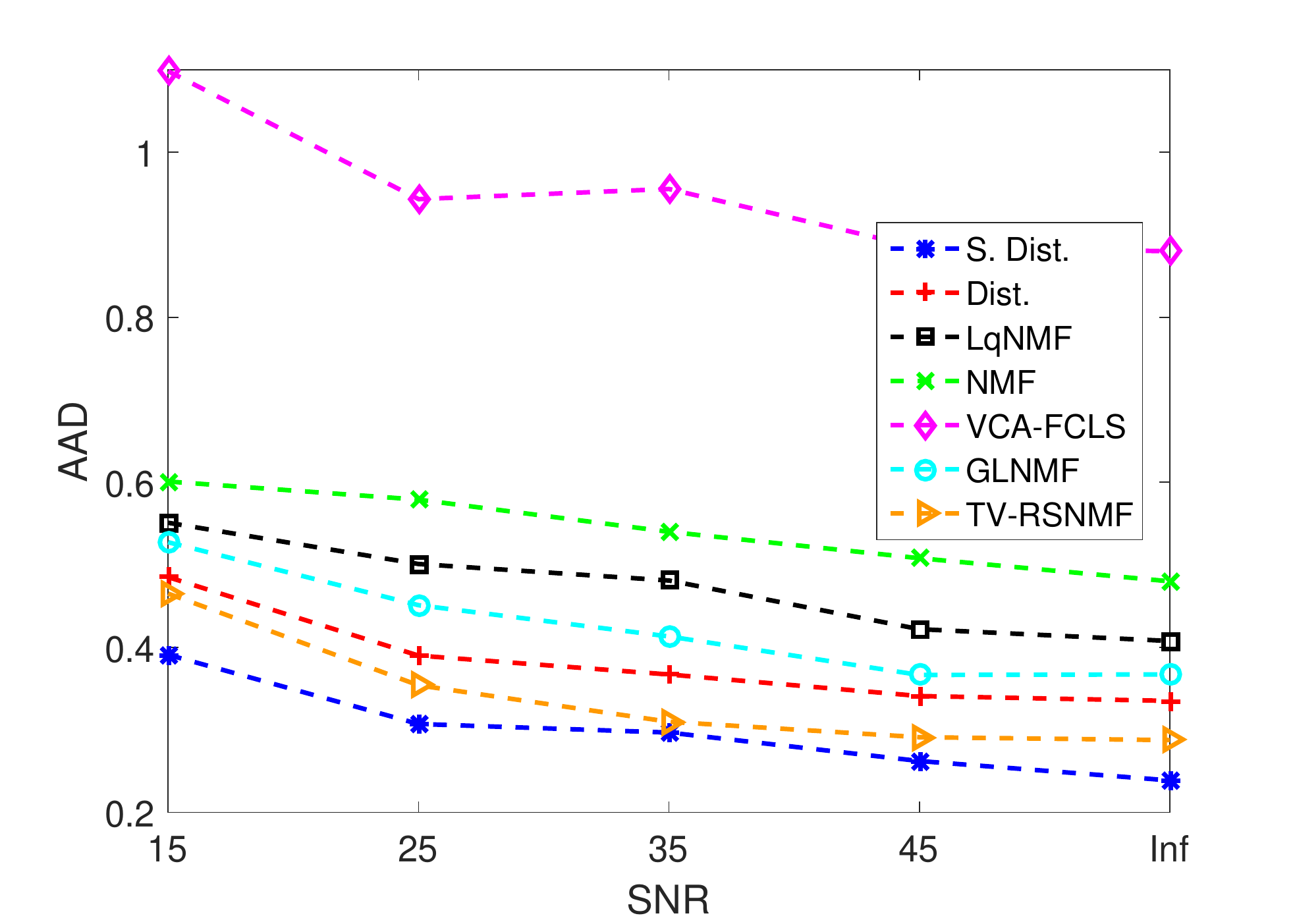}
		\label{fig:s1}
	}
	\subfigure[]{
		\includegraphics[draft=false,width=4cm]{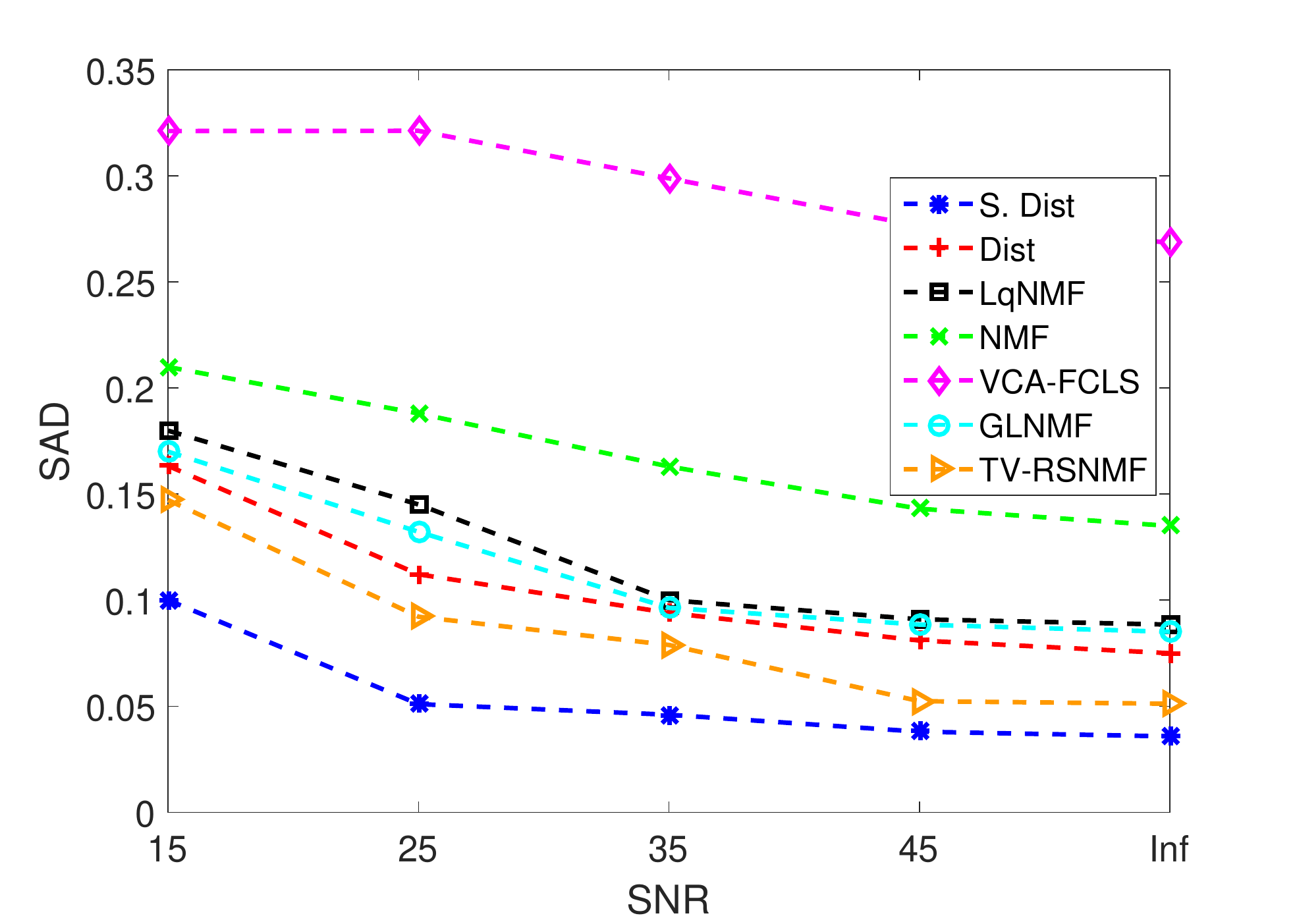}
		\label{fig:s2}
	}
	\caption{(a) The $AAD$ and (b) $SAD$ performance metric of 7 methods versus SNR, using VCA initialization and applied on synthetic data. $SAD$ of the proposed algorithm is star-dashed line with $p=1.75$, $q_1=2$ and $q_2=1$.}
	\label{fig:SADandAAD}
\end{figure}
\begin{figure}
	\begin{center}$
		\begin{array}{lll}
		\includegraphics[draft=false,width=25mm]{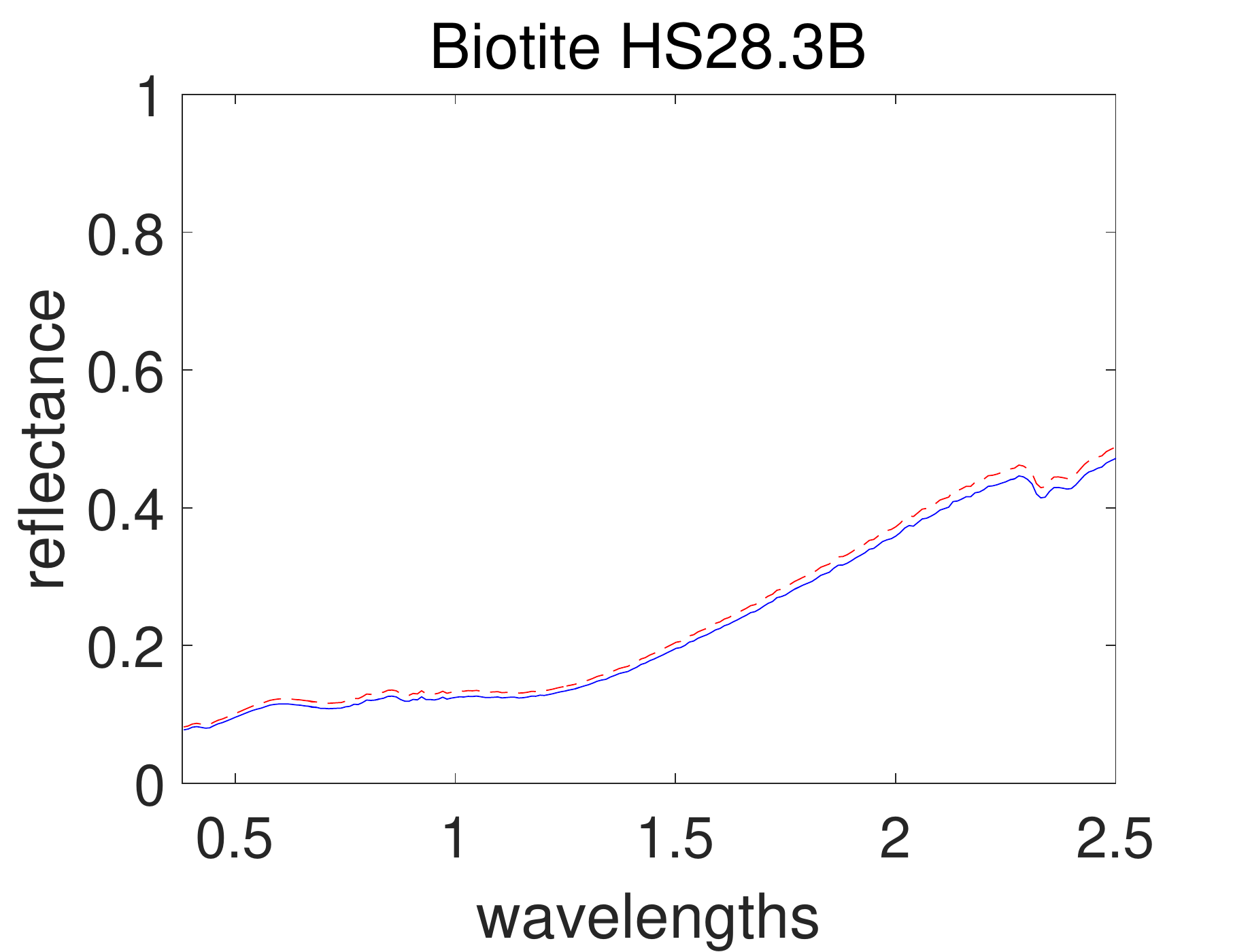}&
		\includegraphics[draft=false,width=25mm]{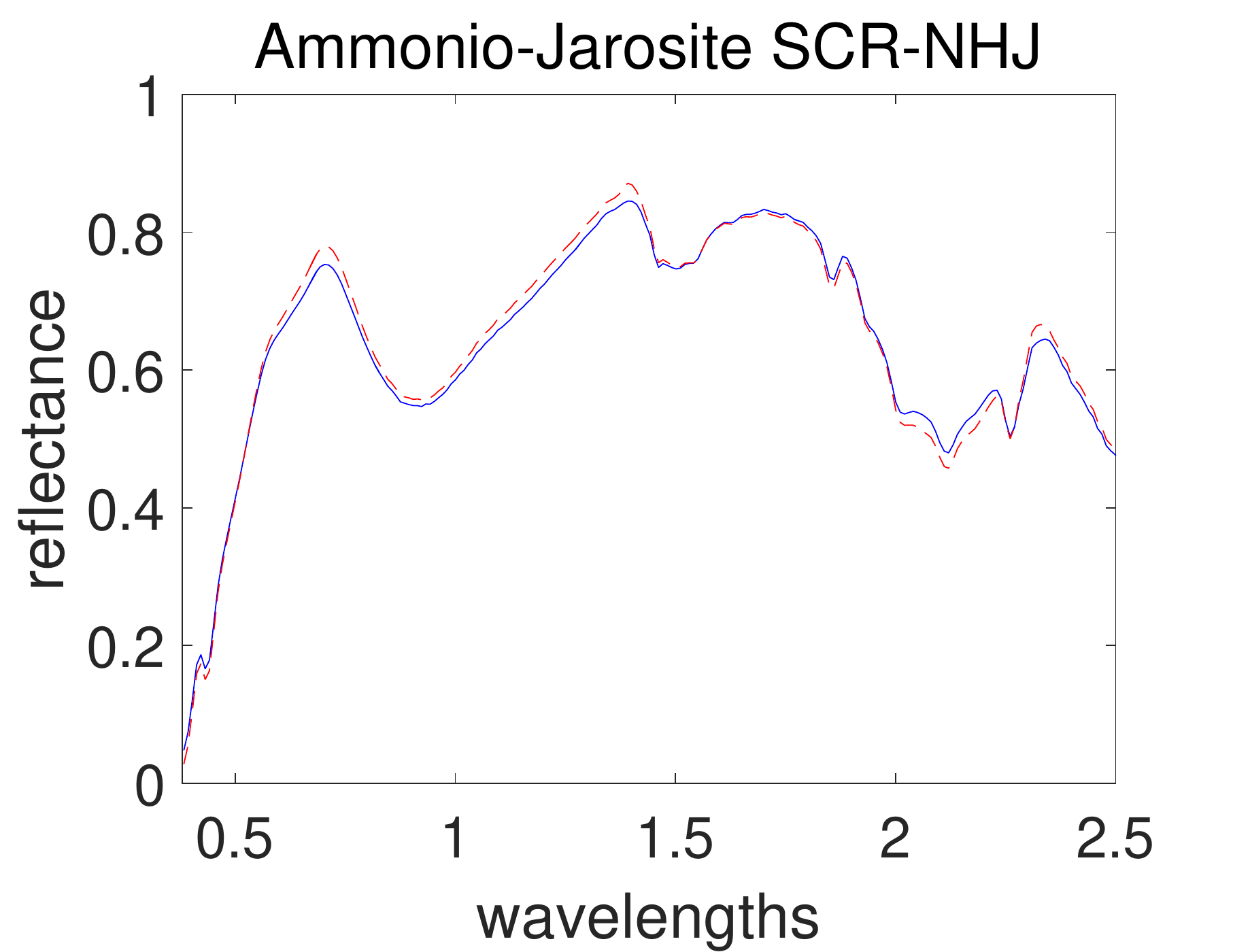}&
		\includegraphics[draft=false,width=25mm]{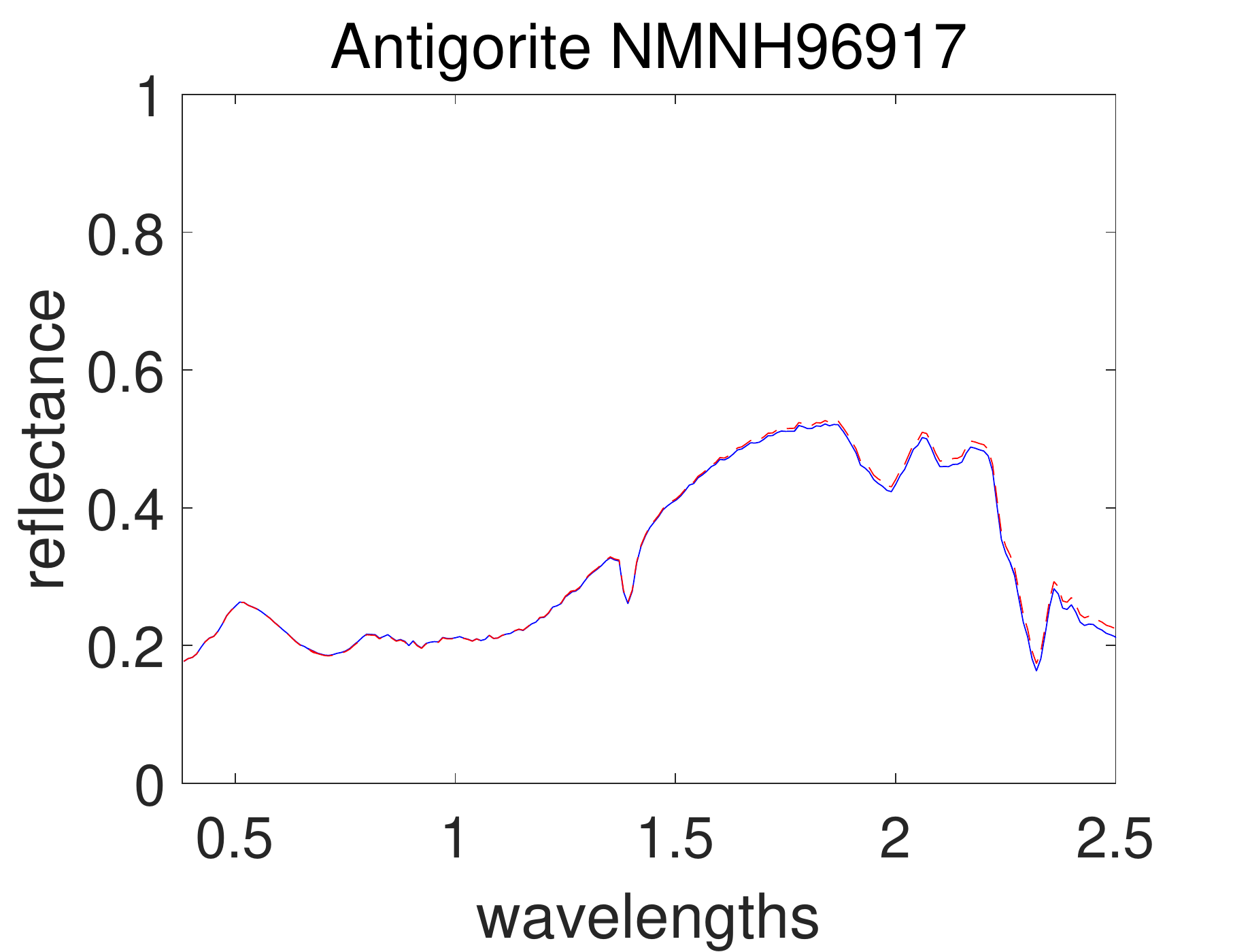}
		\end{array}$
	\end{center}
	
	\begin{center}$
		\begin{array}{lll}
		\includegraphics[draft=false,width=25mm]{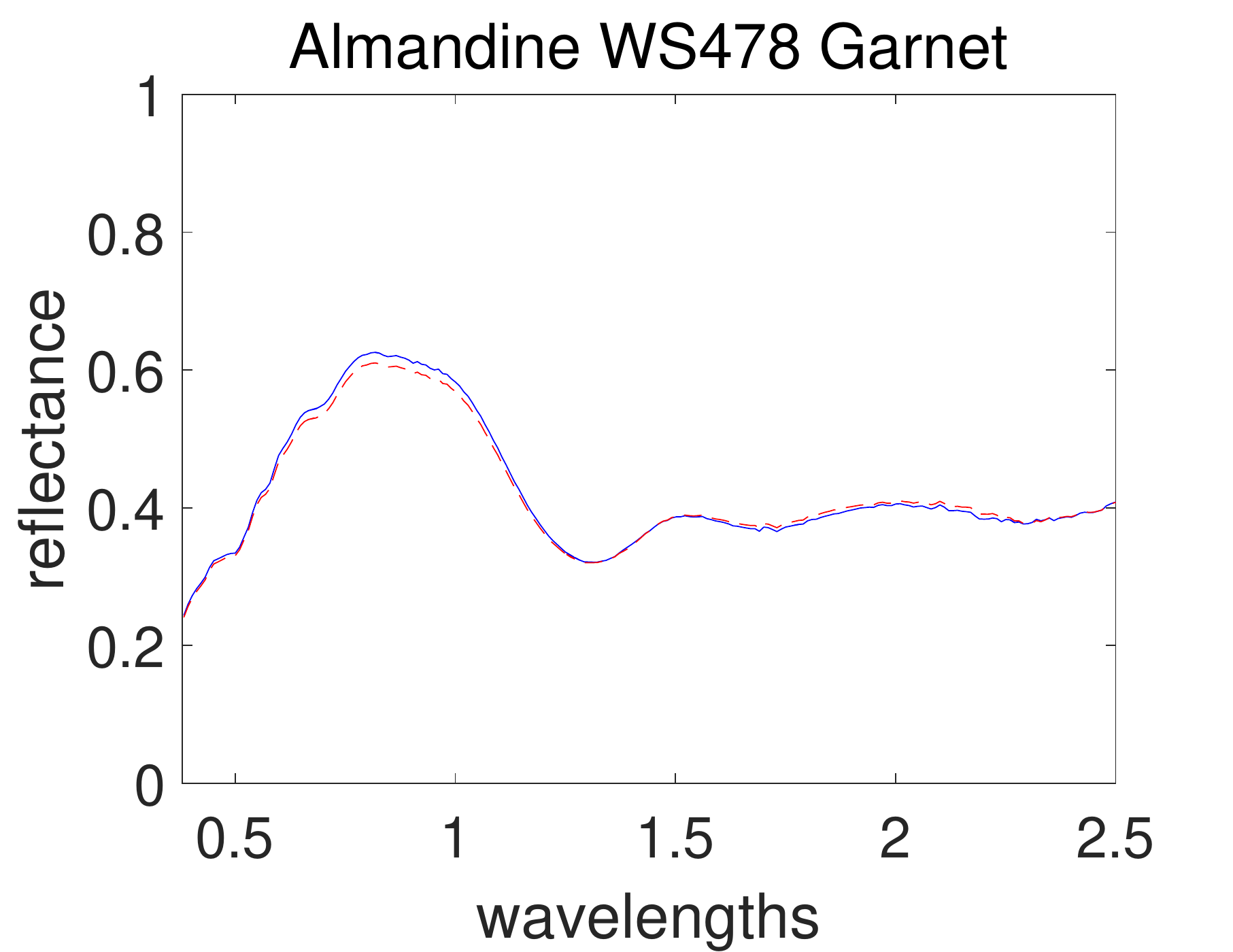}&
		\includegraphics[draft=false,width=25mm]{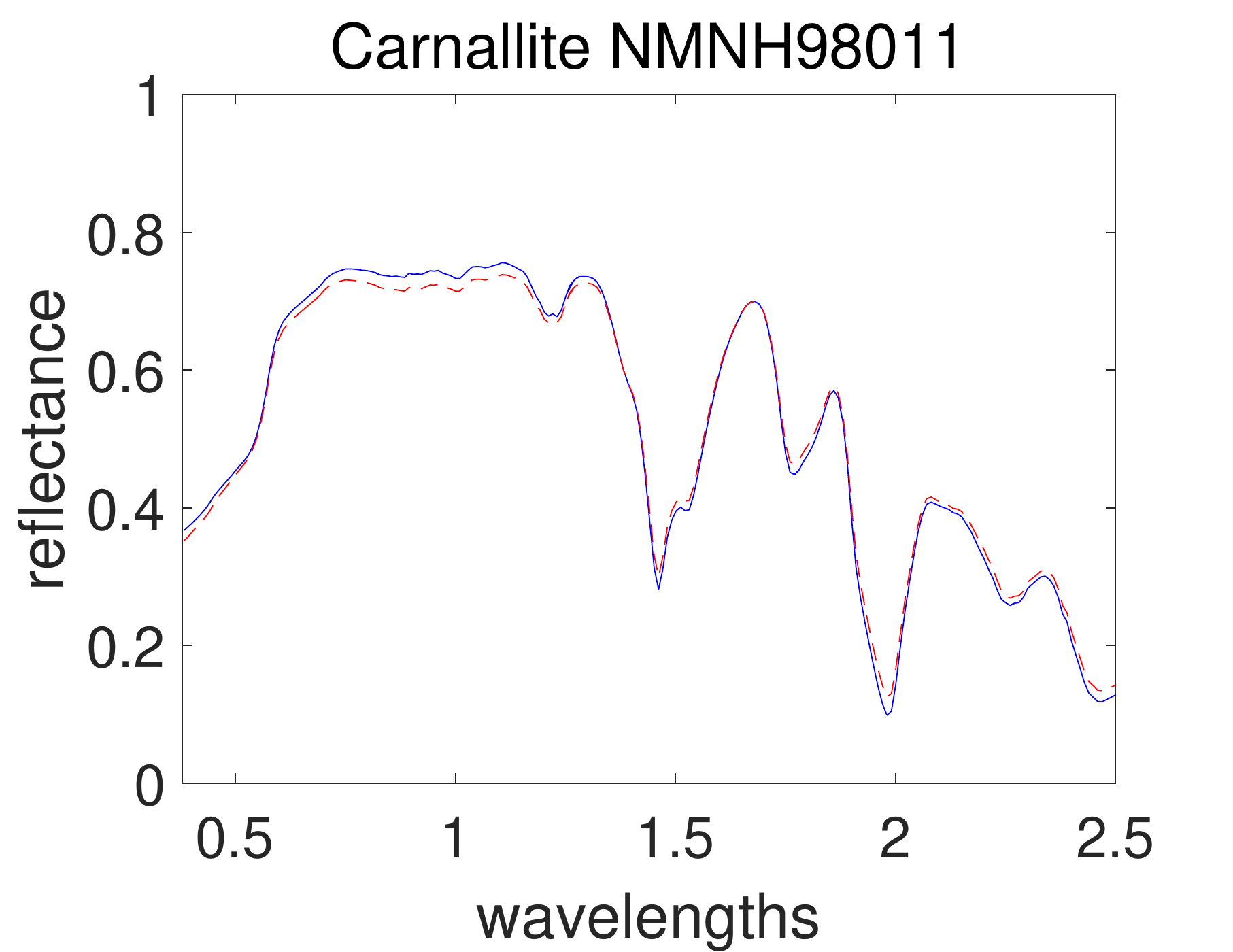}&
		\includegraphics[draft=false,width=25mm]{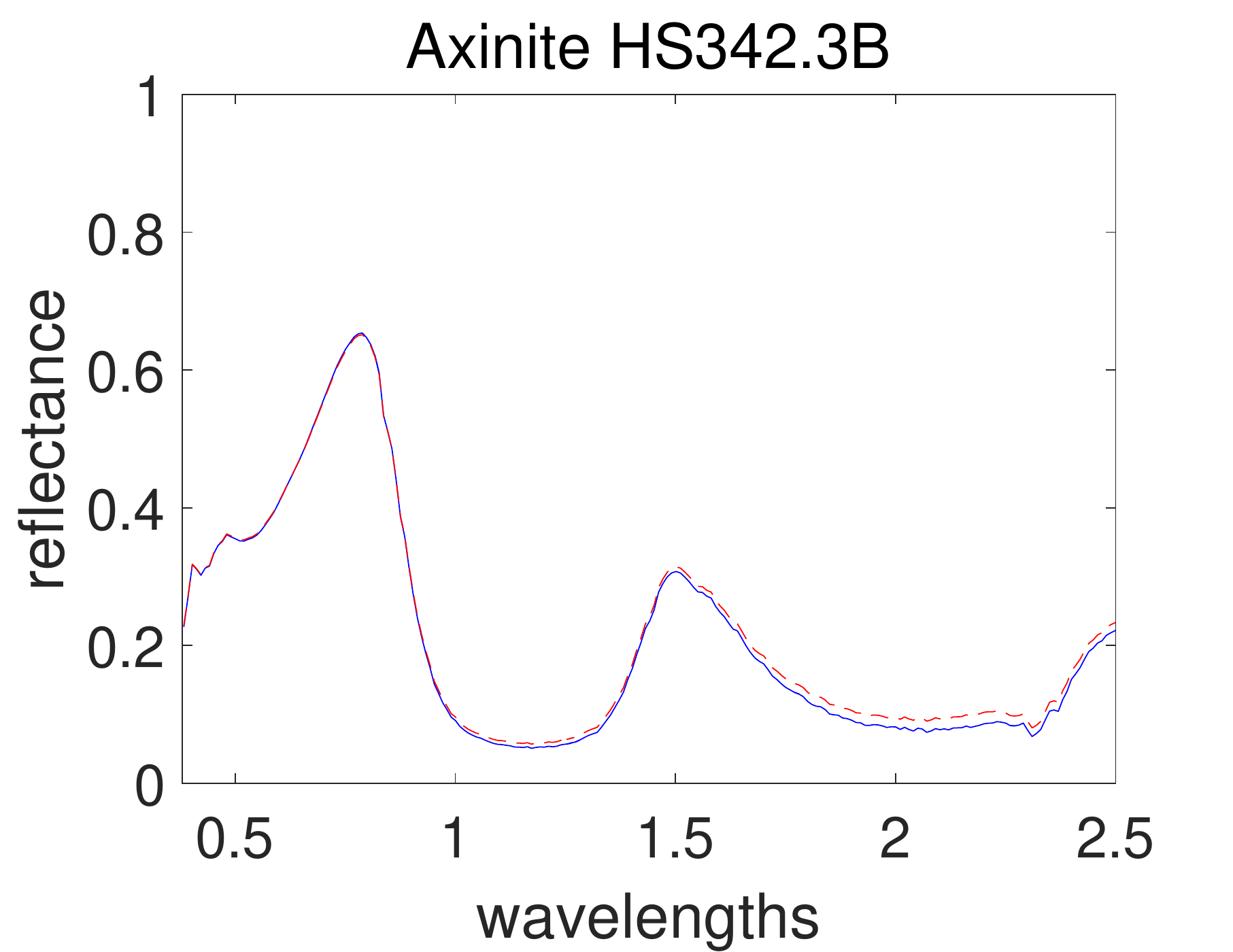}
		\end{array}$
	\end{center}
	\caption{Original spectral signatures (blue solid lines) and estimated signatures of sparsity constrained distributed unmixing algorithm (red dashed lines) versus wavelengths ($\mu m$), on synthetic data and using VCA initialization with SNR=25dB.}
	\label{syn}
\end{figure}
\begin{table}[tb]
	\centering
	\caption{
		Comparison of running time between six algorithms,
		using VCA initialization and SNR$=25dB$.
	}
\scalebox{0.9}{
	\begin{tabular}{c|c}
		\hline
		Methods & Running time (second)\\
		\hline \hline
		NMF &  33.5161\\
		\hline
		$L_{1/2}$-NMF & \textbf{10.8671}\\
		\hline
		GLNMF & 54.6483\\
		\hline
		TV-RSNMF & 40.3321\\
		\hline
		Distributed & 104.5395\\
		\hline
		Sparse. Distributed & \textbf{65.1102}\\
		\hline
	\end{tabular}}
	\label{t1}
\end{table}
\subsection{Experiments on Synthetic Data and Parameter Selection}
First, the proposed algorithm has been applied on synthetic data. to generate this dataset, six signatures of USGS library have been selected randomly, using a 3$\times$3 low pass filter and containing no pure pixels. Then, the zero mean Gaussian noise with seven different levels of SNR has been added to generated data, and performance metrics have been computed by averaging 20 Monte-Carlo runs. To choose the best value of $\mu$ in our experiments, we used the procedure of APPENDIX, and then according to \figurename{~\ref{mu}}, the best value of this parameter has been set equal to 0.02. Also, value of $\eta$ has been considered equal to 0.1 \cite{Chen14}, to gain the best results. In the experiments it is assumed that number of endmembers are preknown, however this parameter can be determined by algorithms like HySime \cite{TFRS08_SubspaceIdentification}.

As the first experiment, \figurename{~\ref{pq}} shows the $SAD$ and $AAD$ performance metrics of proposed method in various $p$, $q_1$ and $q_2$ values versus different levels of SNR. \figurename{~\ref{fig:sub1}} and \ref{fig:sub2} depict that the algorithm with $p=1.75$ in red plus-dashed line eventuates the best result. Values of $p$ have been chosen in accordance with \cite{Wen13}. Also, in \figurename{~\ref{fig:sub3}}, \ref{fig:sub4}, \ref{fig:sub5} and \ref{fig:sub6}, the best values of $q_1$ and $q_2$ are equal to 2. Then the proposed algorithm and some other algorithms such as VCA-FCLS \cite{Nascimento05}, NMF \cite{Lee99}, $L_{1/2}$-NMF \cite{Qian11}, GLNMF \cite{lu13}, TV-RSNMF \cite{he17tvl} and distributed unmixing \cite{Chen14}, have been applied on generated synthetic dataset. The comparison of performance metrics of these seven different methods has been shown in \figurename{~\ref{fig:s1}} and \ref{fig:s2}, where metrics of proposed algorithm with the best results for $p$, $q_1$ and $q_2$ values based on \figurename{~\ref{pq}}, is star-dashed line and excels other methods. The superiority of distributed algorithms in star-dashed and plus-dashed lines, represents preference of using neighborhood information because of correlation between neighboring pixels. \tablename{~\ref{t1}} shows the average of running time of NMF, $L_{1/2}$-NMF, GLNMF, TV-RSNMF, distributed unmixing and proposed method, when $p=1.75$, $q_1=2$ and $q_2=1$ have been chosen. These results have been obtained using MATLAB R2015b with Intel Core i5 CPU at 2.40 GHz and 4 GB memory. This table shows that one of the main advantages of sparse representation is its efficiency and improvement in running time. As the last experiment on synthetic data, \figurename{~\ref{syn}} is illustration of original and estimated spectral signatures for 6 endmembers, when the SNR is set to $25dB$.

\subsection{Experiments on Real Data}
\begin{figure}
	\centering
	\includegraphics[draft=false,width=3in]{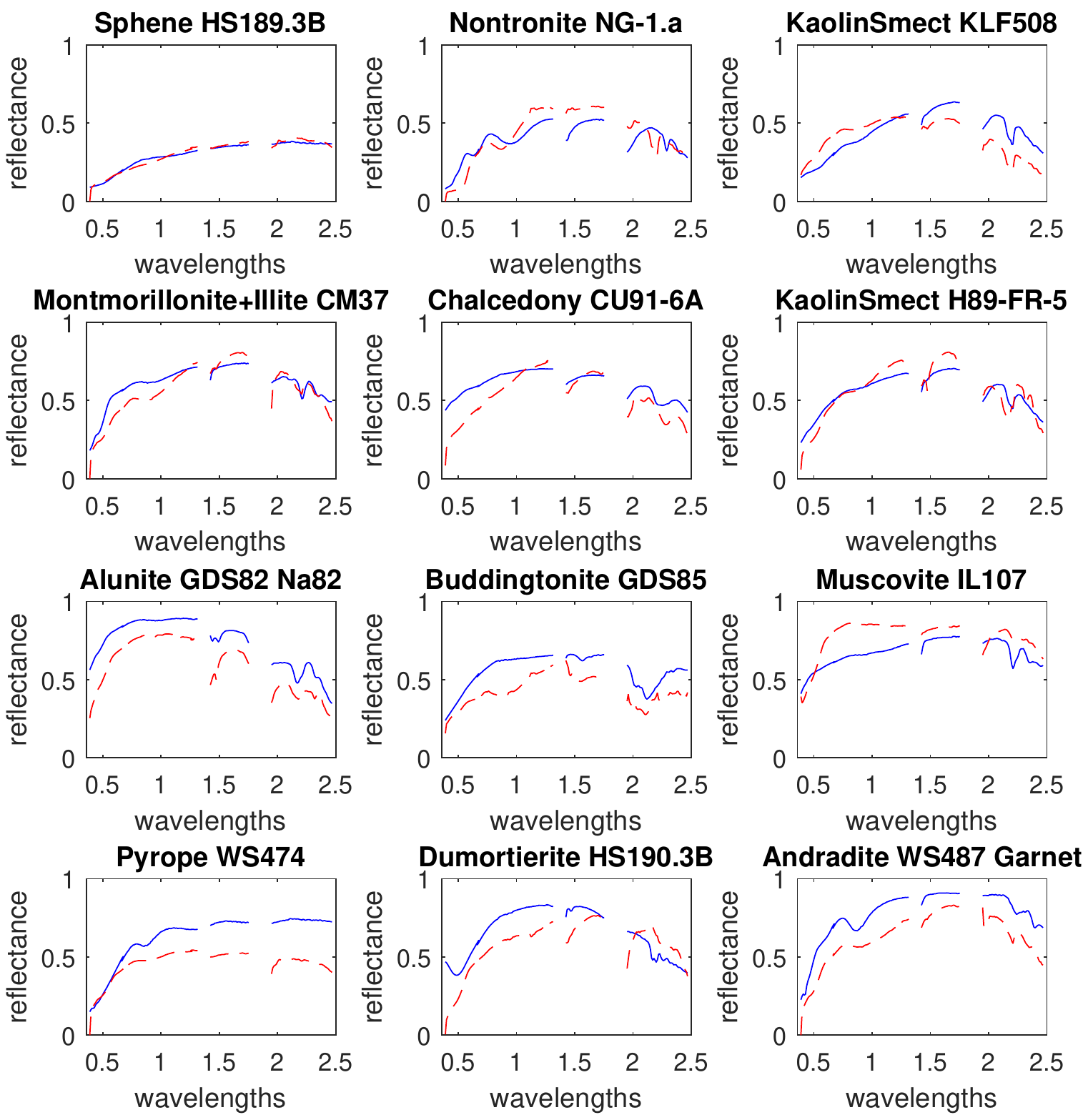}
	\caption{Original spectral signatures (blue solid lines) and estimated signatures of sparsity constrained distributed unmixing (red dashed lines) versus wavelengths ($\mu m$), on AVIRIS Cuprite dataset and using VCA-FCLS initialization.}
	\label{real}
\end{figure}

\begin{figure}[!ht]
	\begin{center}$
		\begin{array}{ll}
		\includegraphics[draft=false,width=27mm]{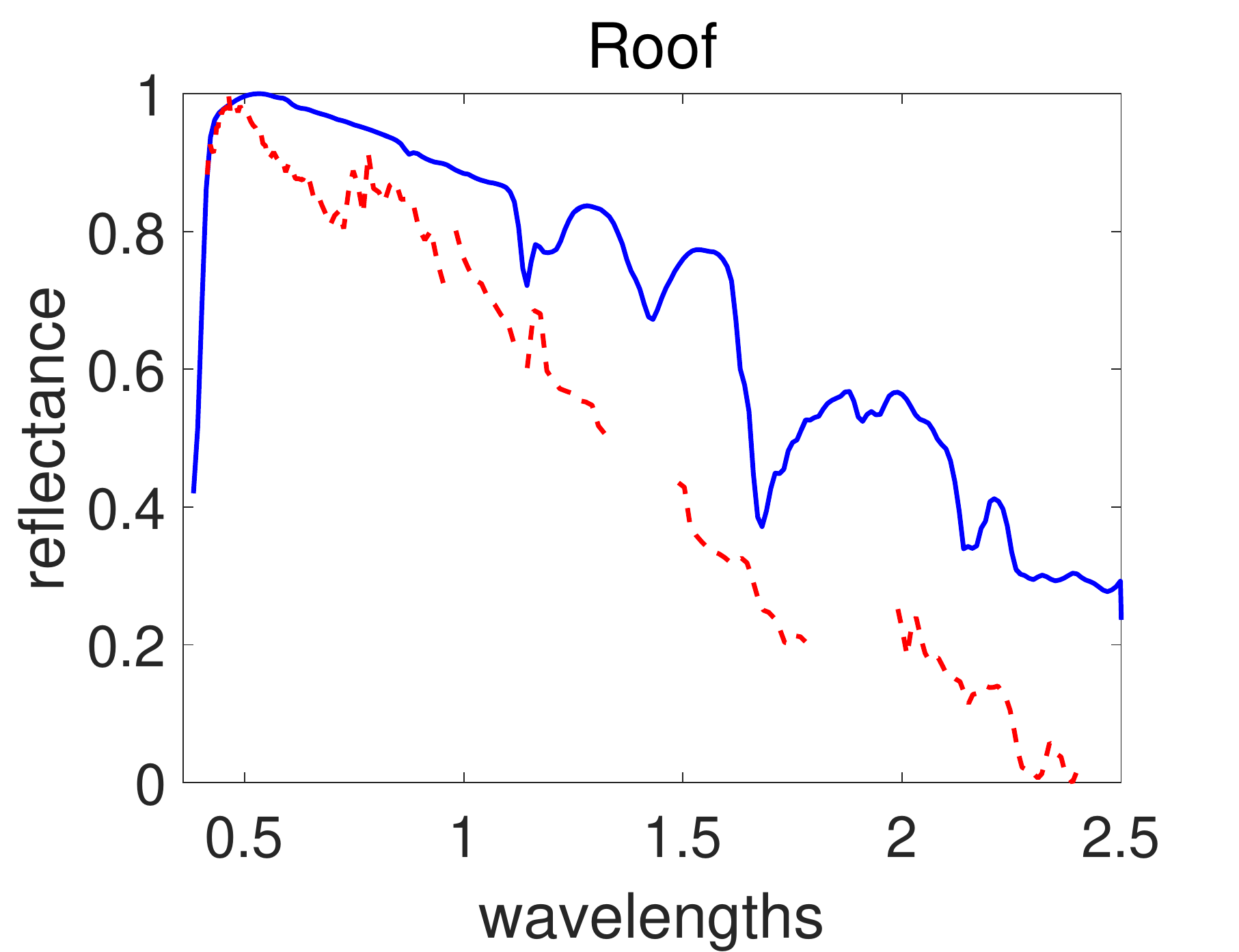}&
		\includegraphics[draft=false,width=27mm]{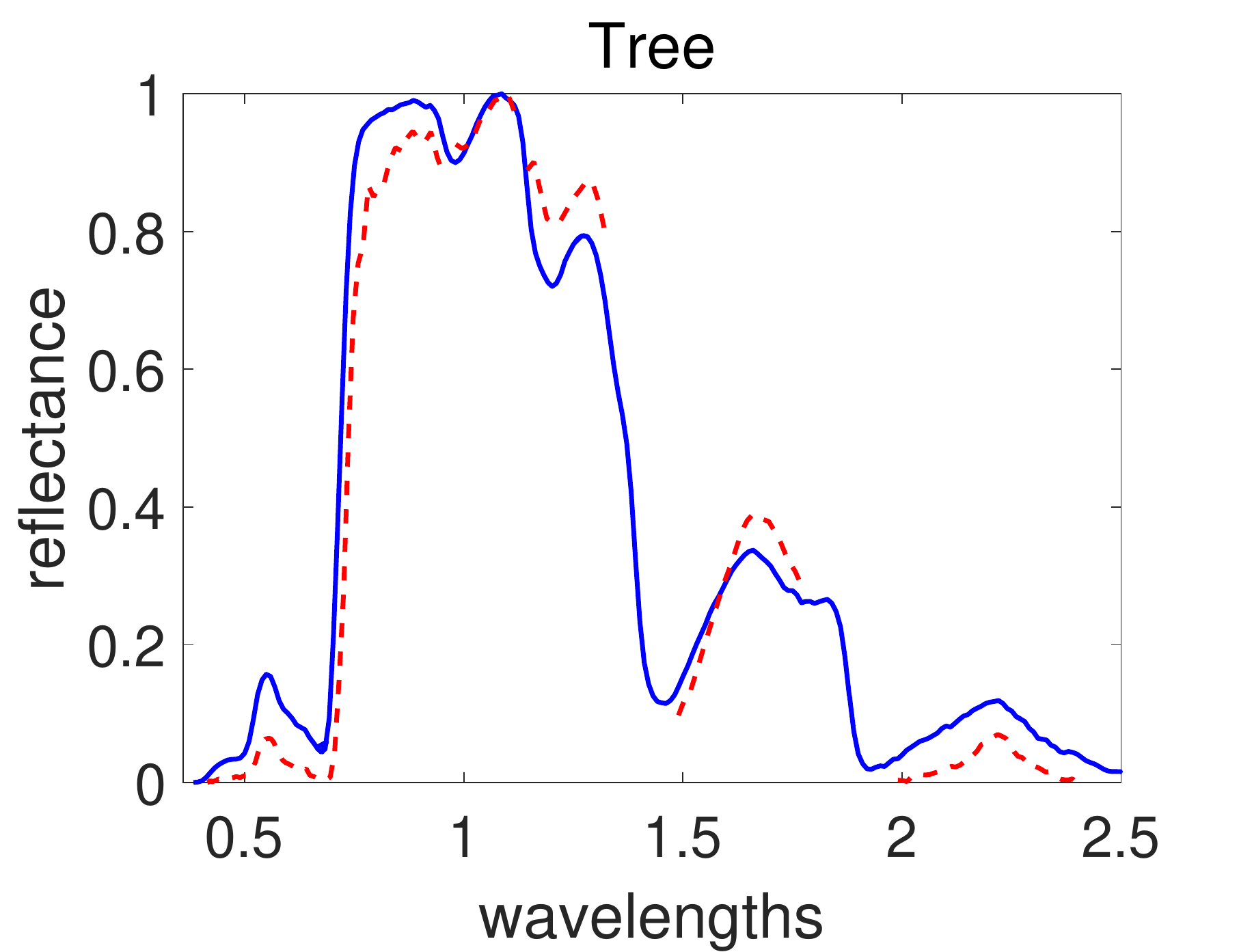}
		\end{array}$
	\end{center}
	
	\begin{center}$
		\begin{array}{ll}
		\includegraphics[draft=false,width=27mm]{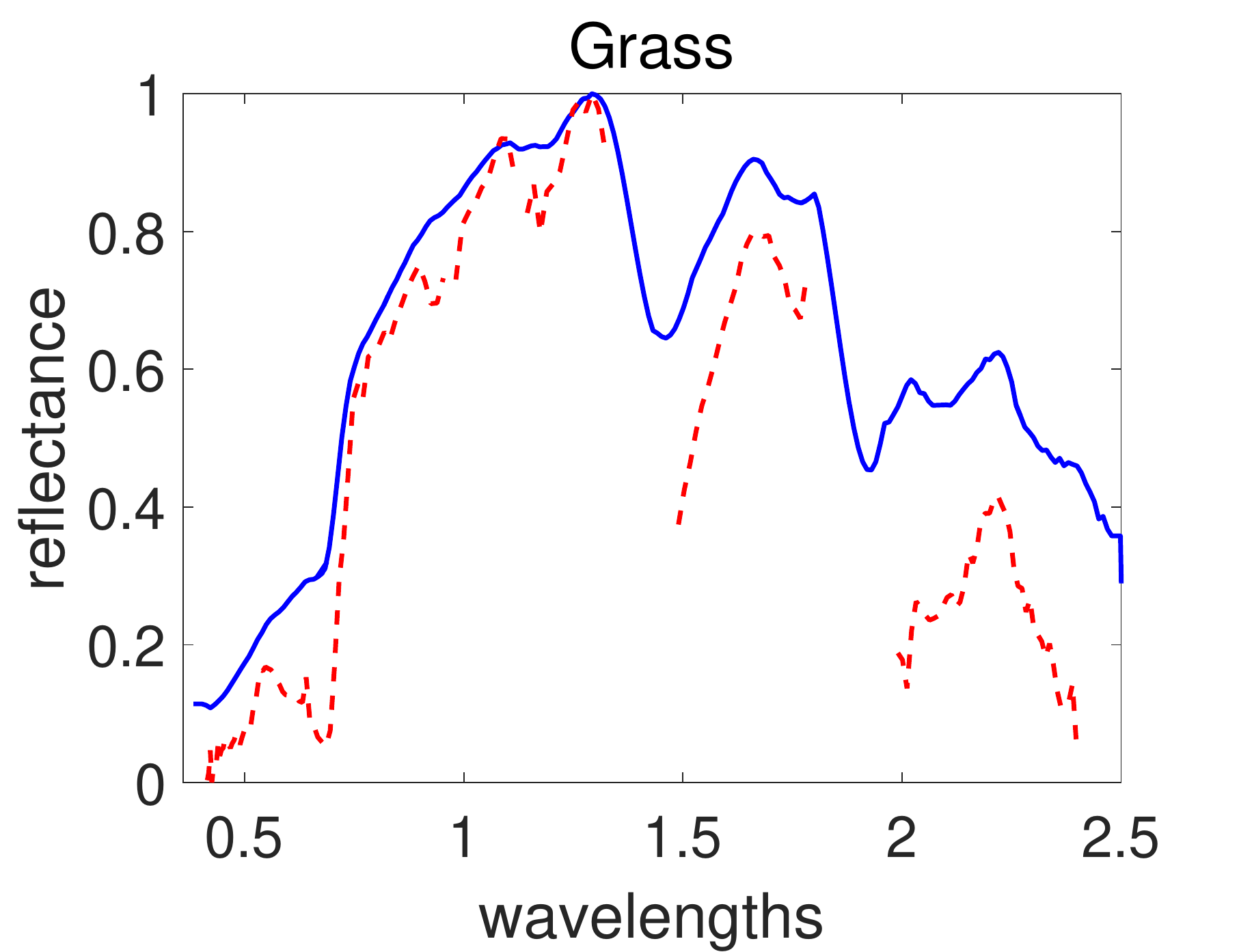}&
		\includegraphics[draft=false,width=27mm]{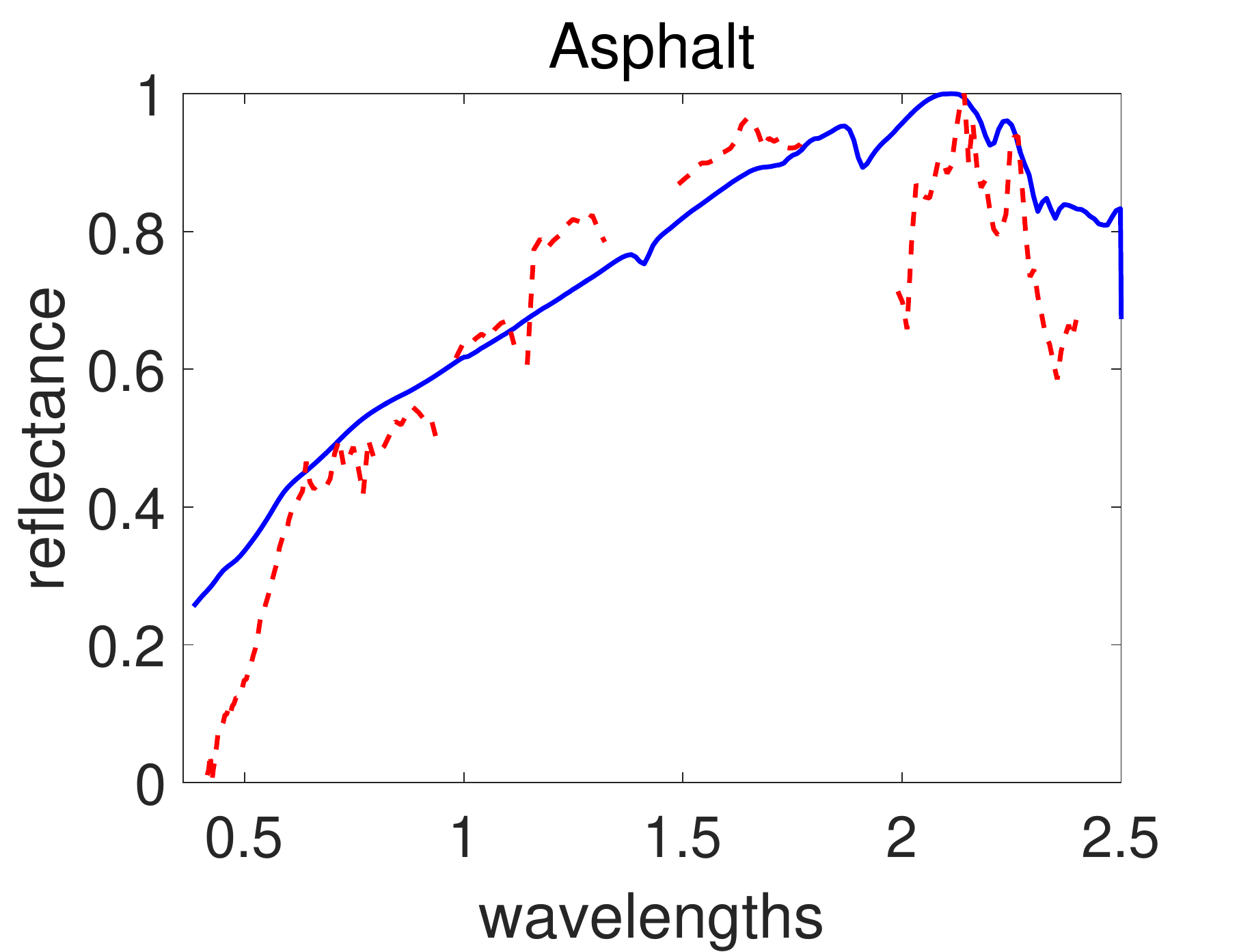}
		\end{array}$
	\end{center}
	\caption{Original spectral signatures (blue solid lines) and estimated signatures of sparsity constrained distributed unmixing (red dashed lines) versus wavelengths ($\mu m$), on HYDICE Urban dataset and using VCA initialization.}
	\label{realhy}
\end{figure}

\begin{figure}[!ht]
	\centering
	\includegraphics[draft=false,width=2.5in]{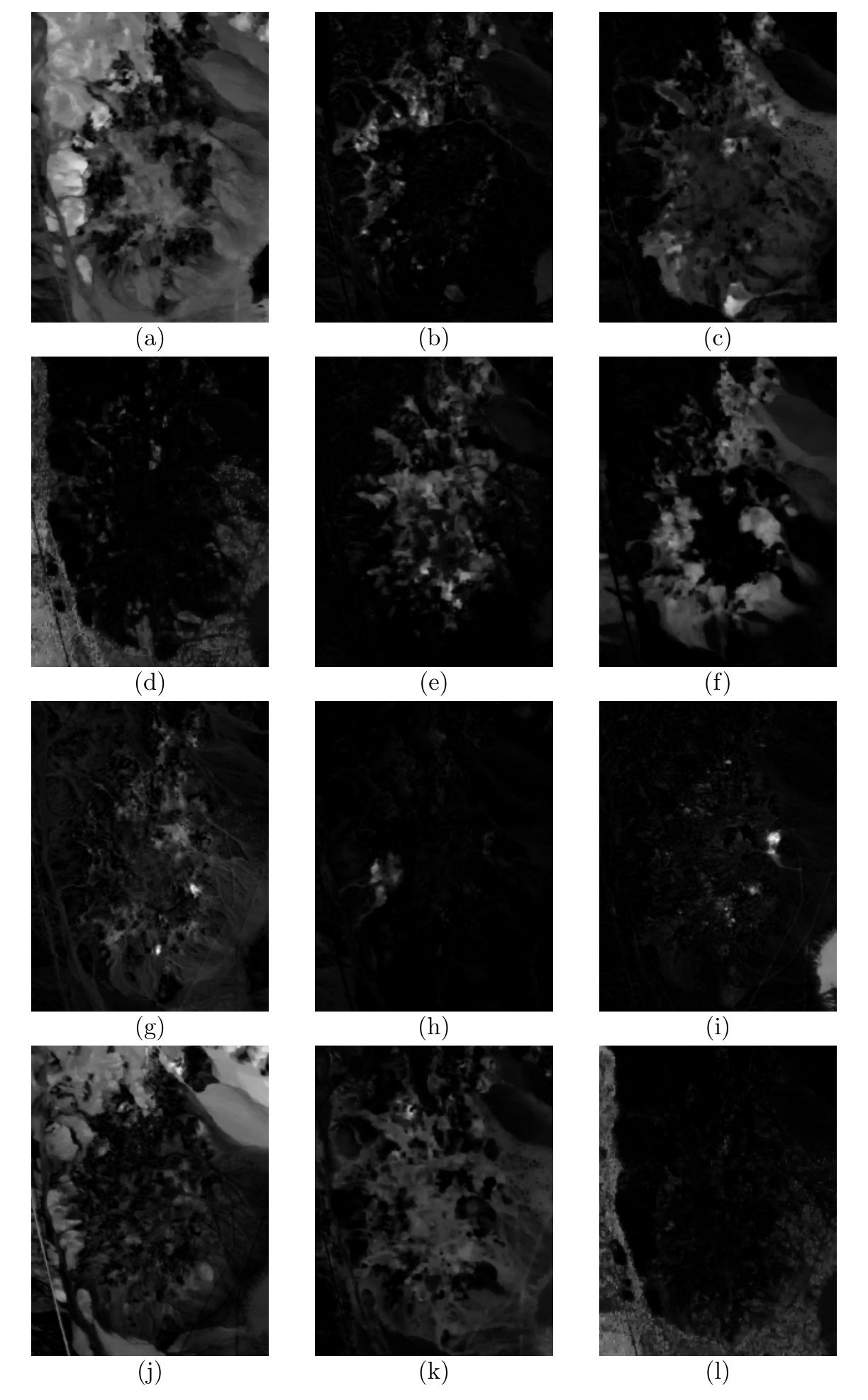}
		\caption{Estimated fractional abundances of AVIRIS Cuprite dataset, using sparsity constrained distributed unmixing. (a) Sphene. (b) Nontronite. (c) Kaolin\_Smect KLF508. (d) Montmorillonite. (e) Chalcedony. (f) Kaolin\_Smect H89-FR-5. (g) Alunite GDS82. (h) Buddingtonite. (i) Muscovite. (j) Pyrope WS474. (k) Dumortierite. (l) Andradite WS487.}
		\label{real1}
\end{figure}

\begin{figure}[!ht]
	\centering
	\includegraphics[draft=false,width=2in]{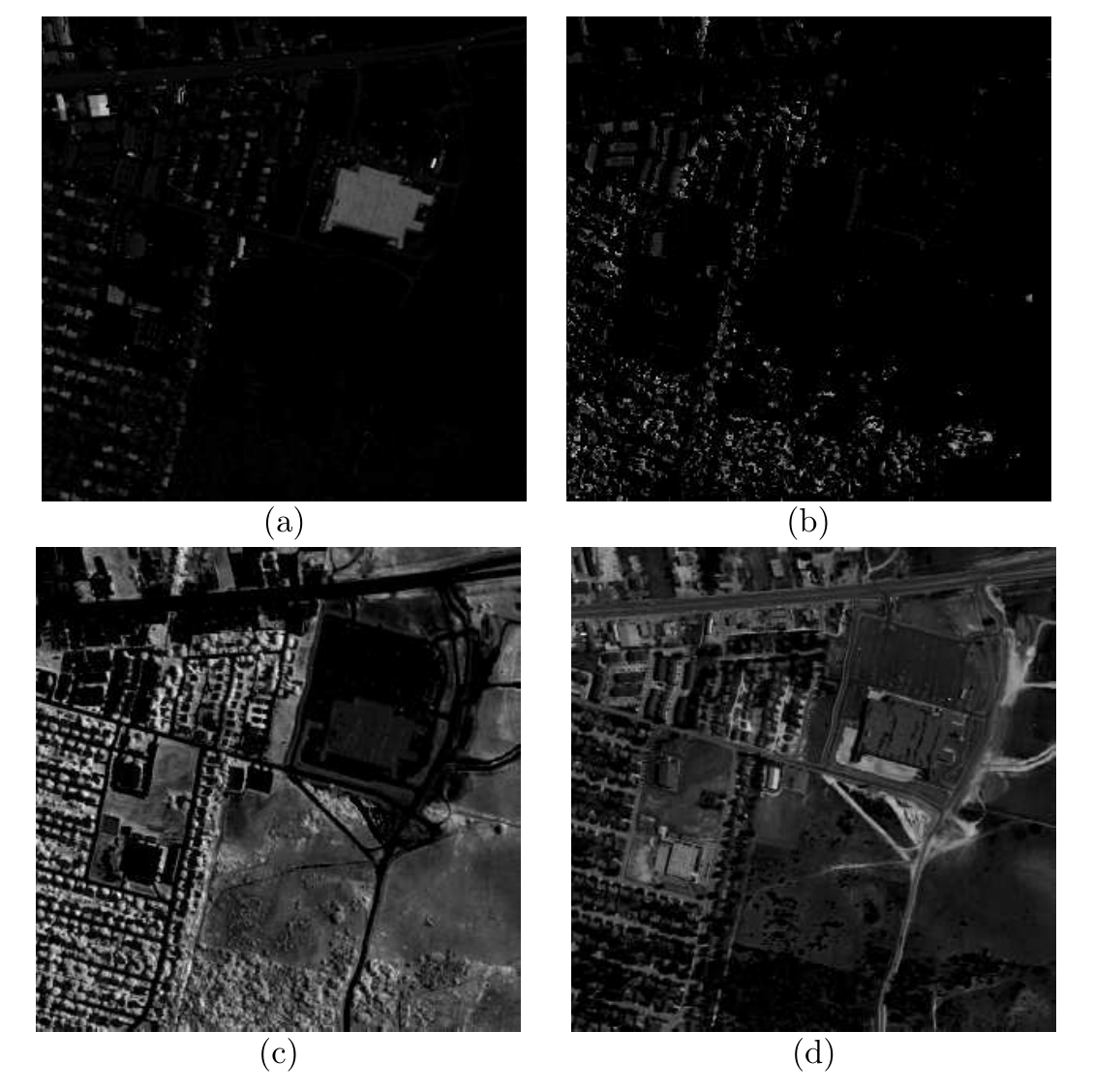}
	\caption{Estimated fractional abundances of HYDICE Urban dataset, using sparsity constrained distributed unmixing. (a) Roof. (b) Tree. (c) Grass. (d) Asphalt.}
	\label{realhyd}
\end{figure}

Afterwards, the proposed algorithm has been applied on AVIRIS Cuprite and HYDICE Urban real datasets described in section III. In the experiments, stopping criteria parameters have been set as follows: the maximum number of iterations is set equal to 200 and the cost function error parameter in (\ref{eq: stop}) has been set to $\epsilon=10^{-8}$. The parameter setting are as follows: $p=2$, $q_1=2$, $q_2=1$, $\mu=0.02$ and $\eta=0.1$. \figurename{~\ref{real}} and {~\ref{realhy}} show simulation results of spectral signatures for AVIRIS and HYDICE datasets respectively. The number of materials in the AVIRIS Cuprite and HYDICE Urban scenes has been considered to be 12 and 4 respectively based on previous works on these datasets. \figurename{~\ref{real1}} and {~\ref{realhyd}} show simulation results of abundance fractions for AVIRIS and HYDICE datasets respectively. To compare quantitatively, $SAD$ performance metric of six related methods applied on these datasets have been summarized in \tablename{~\ref{table1}} and {~\ref{tablehy}} for AVIRIS and HYDICE datasets respectively. In these tables the results of proposed algorithm appear in the last column and show the best $rmsSAD$ value.

\begin{table*}[tb]
	\centering
	\caption{
		The $SAD$ performance metric of six algorithms on AVIRIS Cuprite dataset, using VCA initialization.
	}
	\scalebox{0.9}{\begin{tabular}{c|c|c|c|c|c|c}
		\hline
		materials & VCA-FCLS & $L_{1/2}$-NMF & GLNMF & TV-RSNMF & Distributed & Proposed alg.\\
		\hline \hline
		Sphene &0.3091 &0.2143 & 0.1913 & 0.1583 &0.1561 & \textbf{0.1205} \\
		\hline
		Nontronite & 0.2622 & 0.2518 & 0.1842 & 0.1803 & 0.1944 & \textbf{0.1538}\\
		\hline
		KaolinSmect \#1 & 0.2498 & 0.1653 &0.1638 & 0.1731 & 0.2370 & \textbf{0.1421}\\
		\hline
		Montmorillonite & 0.2609 & 0.2318 & 0.2184 & \textbf{0.2159} & 0.3571 & 0.2163\\
		\hline
		Chalcedony & 0.1934& 0.1995 & 0.1649 & \textbf{0.1588} & 0.1603 & 0.1881\\
		\hline
		KaolinSmect \#2 & 0.3258 &\textbf{0.2542} & 0.2594 & 0.2576 & 0.2873 & 0.2512\\
		\hline
		Alunite &0.3601 &0.3458 & 0.2841 & 0.2551 & 0.3813 & \textbf{0.2158}\\
		\hline
		Buddingtonite &0.2402 & \textbf{0.1693} & 0.2068 &0.2034 & 0.2514 & 0.1778\\
		\hline
		Muscovite & 0.3917 & 0.1584 & \textbf{0.1471} & 0.1563 & 0.4682 & 0.1826\\
		\hline
		Pyrope &0.2851 &0.3361 & 0.3148 & 0.2392 & \textbf{0.2132} & 0.2385 \\
		\hline
		Dumortierite & \textbf{0.2311} & 0.2453 & 0.2632 & 0.2686 & 0.3381 & 0.2917 \\
		\hline
		Andradite & 0.4492 & 0.3829 & 0.3021 & 0.3136 & 0.3711 & \textbf{0.2963}\\
		\hline\hline
		rmsSAD &0.3049 & 0.2562 & 0.2317 & 0.2207 & 0.2998 & \textbf{0.2131}\\
		\hline
	\end{tabular}}
	\label{table1}
\end{table*}
\begin{table*}[!t]
	\centering
	\caption{
		The $SAD$ performance metric of six algorithms on HYDICE Urban dataset, using VCA initialization.
	}
	\scalebox{0.9}{\begin{tabular}{c|c|c|c|c|c|c}
		\hline
		materials & VCA-FCLS&$L_{1/2}$-NMF & GLNMF & TV-RSNMF & Distributed & Proposed alg.\\
		\hline \hline
		Roof &0.4671 & 0.3461 & 0.3486 & 0.3327 & 0.3831 & \textbf{0.3294} \\
		\hline
		Tree & 0.2711 & \textbf{0.1492} & 0.1673 & 0.1572 & 0.2052 & 0.1521\\
		\hline
		Asphalt & 0.3077 & 0.2984 & 0.2096 & \textbf{0.2054} & 0.2469 & 0.2118\\
		\hline
		Grass & 0.2089 & 0.1461 & 0.1283 & 0.1249 & 0.1344 & \textbf{0.1019}\\
		\hline\hline
		rmsSAD &0.3279 & 0.2512 & 0.2291 & 0.2198 & 0.2588 & \textbf{0.2161}\\
		\hline
	\end{tabular}}
	\label{tablehy}
\end{table*}

\section{Conclusion and Future Work}
Hyperspectral remote sensing is a prominent research topic in data processing. The purpose of spectral unmixing is decomposition of pixels in the scene into their constituent materials. This paper used the sparsity constrained distributed unmixing method that improved estimation of spectral signature of endmembers and their abundances. This new algorithm considered sparsity and neighborhood information. In our experiments, the best power of LMP and $L_q$ norms have been found, using AAD and SAD performance metrics. Simulation results on synthetic and real datasets illustrated better performance of proposed approach compared with VCA-FCLS, NMF, $L_{1/2}$-NMF, GLNMF, TV-RSNMF and Distributed Unmixing methods. Furthermore, the algorithm of this paper achieved faster convergence in comparison with the distributed method. Also by obtaining optimum $p$ and $q_1$ values and adding optimum norm of sparsity constraint, the proposed method gained about 25\% and 79\% improvement of SAD and AAD respectively, in SNR$=$25 toward distributed method. In this paper, the neighborhood information has been used as spatial information, however spectral information can be useful in the SU problem. Therefore, using clustering algorithms as a preprocessing step, and then adopting the clustered multitask network model for the distributed unmixing method, is expected to eventuate better results. Additionally, recently proposed multilayer or deep NMF structures can also be used in conjunction with the proposed method to gain better results.

\section*{Acknowledgment}
The authors would like to acknowledge editor, associate editor, and five anonymous reviewers for their helpful comments, and Dr. Mehdi Korki for his useful suggestions to improve the revised paper.

\appendix
\section*{Mean Error Convergence Analysis}
Here the stochastic behavior of the proposed algorithm is studied. First, the weight error vector is defined as $\mathbf{\nu}_k=\mathbf{s}_k-\mathbf{s}^*$, where $\mathbf{s}^*$ is optimum abundance vector \cite{farhang13}. We collect information from all pixels and these vectors are stacked by $col\{.\}$ as follows
$\mathbf{\nu}=col\{\mathbf{\nu}_k\}_{k=1}^N$, $\mathbf{s}=col\{\mathbf{s}_k\}_{k=1}^N$ and $\mathbf{s}^*=col\{\mathbf{s}_k^*\}_{k=1}^N$.

Then we can write the following equation using these expressions and the recursive equation of (\ref{eq: 19}) with constant values of $p$, $q_1$ and $q_2$, that is chosen equal to 2, 1 and 1, respectively:
\begin{equation}
\begin{aligned}
\label{eq: nu1}
&\mathbf{\nu}^{i+1}=\mathbf{\nu}^i-\mu \mathbf{H}_a^i \mathbf{\nu}^i+\mu \mathbf{r}_{\mathbf{ya}}^i &\\
&-\mu \eta \mathbf{Q}(\mathbf{\nu}^i+\mathbf{s}^*)-\mu\lambda (\mathbf{\nu}^i+\mathbf{s}^*)&
\end{aligned}
\end{equation}
where we can define $\mathbf{Q}=\mathbf{I}_{LN}-\mathbf{P}\otimes\mathbf{I}_L$ \cite{Chen14} that $\otimes$ is the Kronecker product \cite{Kronecker04}, and $\mathbf{P}$ is a $N\times N$ matrix with $\rho_{kl}$ components, $\mathbf{H}_{\mathbf{a}}$ is a $LN\times LN$ diagonal matrix defined as $\mathbf{H}_{\mathbf{a}}=diag\{\mathbf{a}_k\mathbf{a}_k^T\}_{k=1}^N$
and $\mathbf{r}_{\mathbf{ya}}$ is a $LN\times 1$ vector $\mathbf{r}_{\mathbf{ya}}=col\{\mathbf{a}_k^T\mathbf{y}_k\}_{k=1}^N$.

Now by taking expectation from both side of (\ref{eq: nu1}), the following equation is obtained:
\begin{equation}
\begin{aligned}
\label{eq: nu2}
&\mathbb{E}\{\mathbf{\nu}^{i+1}\}=\mathbb{E}\{\mathbf{\nu}^i\}\big[\mathbf{I}_{LN}-\mu (\mathbb{E}\{\mathbf{H}_\mathbf{a}\}+\eta\mathbf{Q}+\lambda)\big]&\\
&-\mu \eta \mathbf{Q}\mathbf{s}^*-\mu\lambda\mathbf{s}^*&
\end{aligned}
\end{equation}
where $\mathbb{E}\{\mathbf{H}_\mathbf{a}\}$ is taken as $\mathbb{E}\{\mathbf{H}_\mathbf{a}\}=diag\{\mathbf{R}_1,...,\mathbf{R}_N\}$ with $\mathbf{R}_k=\sum\limits_{l\in \mathcal{N}_k^-} \mathbf{R}_{a,l}$ and $\mathbf{R}_{a,k}=\mathbb{E}\{a_k a_k^T\}>0$.

Then the proposed algorithm converges, if the following inequality is satisfied \cite{S13}:
\begin{equation}
\label{eq:r}
\rho\big(\mathbf{I}_{LN}-\mu (\mathbb{E}\{\mathbf{H}_\mathbf{a}\}+\eta\mathbf{Q}+\lambda)\big)<1
\end{equation}
where $\rho(.)$ is spectral radius of its matrix argument, and according to the following definition of block maximum norm for any vector \cite{S13}, $||x||_{b,\infty}\triangleq \max \limits_{1\leq k\leq N} ||x_k||$ and its properties, we get:
\begin{equation}
\begin{aligned}
\label{eq:rr}
&\rho\big(\mathbf{I}_{LN}-\mu (\mathbb{E}\{\mathbf{H}_\mathbf{a}\}+\eta\mathbf{Q}+\lambda)\big)&\\
&\leq  ||\mathbf{I}_{LN}-\mu (\mathbb{E}\{\mathbf{H}_\mathbf{a}\}+\eta\mathbf{Q}+\lambda)||_{b,\infty}&\\
&\leq ||\mathbf{I}_{LN}-\mu \mathbb{E}\{\mathbf{H}_\mathbf{a}\}-\mu\lambda-\mu \eta \mathbf{I}_{LN}||_{b,\infty}&\\
& +\mu \eta ||\mathbf{P}\otimes\mathbf{I}_L||_{b,\infty}&
\end{aligned}
\end{equation}
Since $\mathbf{P}$ is a right stochastic matrix \cite{Chen14}, we have $||\mathbf{P}\otimes\mathbf{I}_L||_{b,\infty}=1$.

So, the inequality of (\ref{eq:r}) is simplified to:
\begin{equation}
\label{eq:r1}
\rho\big((1-\mu \eta)\mathbf{I}_{LN}-\mu \mathbb{E}\{\mathbf{H}_\mathbf{a}\}\big)-\mu \lambda+\mu \eta<1
\end{equation}
Note that, the spectral radius is used, because its argument is a diagonal Hermitian matrix. Therefore upper bound of $\mu$ is obtained as:
\begin{equation}
\label{eq:mio}
0<\mu<\frac{2}{max_k\{\lambda_{max}(\mathbf{R}_k)\}+2\eta-\lambda}
\end{equation}
where $\lambda_{max}(.)$ is the maximum eigenvalue of its argument. Afterwards this upper bound is achieved equal to 0.1652, using the synthetic dataset in our experiments.
\ifCLASSOPTIONcaptionsoff
  \newpage
\fi



%

\bibliographystyle{ieeetr}
\bibliography{bare_jrnl}

%









\end{document}